\documentclass{article}
\usepackage{PRIMEarxiv}
\usepackage[utf8]{inputenc} 
\usepackage[T1]{fontenc}    
\usepackage{hyperref}       
\usepackage{url}            
\usepackage{booktabs}       
\usepackage{amsfonts}       
\usepackage{nicefrac}       
\usepackage{microtype}      
\usepackage{lipsum}
\usepackage{fancyhdr}       
\usepackage{graphicx}       
\usepackage{amsmath,amsfonts}
\usepackage{booktabs}
\usepackage{enumerate}
\usepackage{bm}
\usepackage{xcolor}
\usepackage{adjustbox}
\usepackage{multirow}
\graphicspath{{media/}}     

\title{Decoding for Punctured Convolutional and Turbo Codes: A Deep Learning Solution for Protocols Compliance}

\author{
  Yongli Yan \\
  Department of Electronic Engineering \\
  Tsinghua University \\
  \texttt{yanyongli@tsinghua.edu.cn} \\
  \And
  Linglong Dai \\
  Department of Electronic Engineering \\
  Tsinghua University \\
  \texttt{daill@tsinghua.edu.cn} \\
}

\newcounter{magicrownumbers}

\begin{document}
\maketitle

\begin{abstract}
Neural network-based decoding methods show promise in enhancing error correction performance but face challenges with punctured codes. In particular, existing methods struggle to adapt to variable code rates or meet protocol compatibility requirements. This paper proposes a unified long short-term memory (LSTM)-based neural decoder for punctured convolutional and Turbo codes to address these challenges. The key component of the proposed LSTM-based neural decoder is puncturing-aware embedding, which integrates puncturing patterns directly into the neural network to enable seamless adaptation to different code rates. Moreover, a balanced bit error rate training strategy is designed to ensure the decoder’s robustness across various code lengths, rates, and channels. In this way, the protocol compatibility requirement can be realized. Extensive simulations in both additive white Gaussian noise (AWGN) and Rayleigh fading channels demonstrate that the proposed neural decoder outperforms conventional decoding techniques, offering significant improvements in decoding accuracy and robustness.
\end{abstract}

\section{Introduction}
Artificial intelligence (AI) has demonstrated exceptional capabilities in various fields, such as natural language processing (e.g., ChatGPT)~\cite{chatgpt} and computer vision~\cite{ml_for_cv}, driving breakthroughs in applications previously dominated by hand-engineered methods. These advances in AI have not only revolutionized traditional domains but also sparked interest in its application to the evolving telecommunications industry. With the ongoing transition towards 6th generation (6G) networks, AI’s nonlinear modeling capabilities are being explored to enhance wireless communication systems. Recognizing its potential, the 3rd generation partnership project (3GPP) has established the AI-for-RAN working group in 2022~\cite{3gpp_tr37817}, focusing on incorporating AI into the radio access network (RAN) to improve key performance metrics such as efficiency and capacity.

\subsection{Prior Works}
Building on AI’s potential in telecommunications, its application in physical layer signal processing for wireless communications has garnered significant attention, particularly in tasks such as channel estimation, signal detection, and channel decoding~\cite{ml_ch_est0, ml_ch_est1, ml_mimo_det0, ml_mimo_det1, ml_ch_dec0, ml_ch_dec1}, where it demonstrates notable advantages. In contrast, traditional signal processing algorithms have typically been implemented on central processing units (CPUs), digital signal processors (DSPs), or application-specific integrated circuits (ASICs), relying on \textbf{\textit{serial processing}}. However, due to the highly \textbf{\textit{parallel nature}} of AI algorithms, traditional serial processing architectures struggle to meet their demands, thus driving the shift toward graphics processing units (GPUs), which are optimized for parallel computation and capable of efficiently handling large-scale, simultaneous operations.

As a pioneer in this shift and a leading designer of GPUs, NVIDIA has restructured cellular wireless network receivers using AI. For instance, NVIDIA developed the Sionna signal processing AI library~\cite{sionna_lib} and used it to create multi-user, real-time neural network (NN) receivers compatible with the 5th generation new radio (5G NR) protocol~\cite{mumimo_5gnr_sionna, std_5gnr_sionna}. While AI has been successfully applied to channel estimation, signal detection, and demodulation in NVIDIA’s work, integrating neural network-based decoders into the receiver presents significant challenges. One major issue is the generalization of neural network decoders, particularly when applied to varying code rates, which can limit their performance and flexibility in diverse scenarios~\cite{6g_unify_decoder}.

Puncturing, which discards part of the encoded data to form different code rates and improve spectral efficiency, is essential in real-world wireless communication systems. Control channels typically employ lower code rates for high reliability, while data channels use more flexible and higher code rates to accommodate diverse transmission conditions and maximize throughput. Linear block codes (e.g., low-density parity-check (LDPC) and Polar codes) and sequential codes (e.g., convolutional and Turbo codes) are widely used in commercial communication protocols~\cite{ldpc_code, polar_code, conv_code, turbo_code}. For example, Wi-Fi protocols~\cite{ieee_80211_std} support four code rates for convolutional and LDPC codes, while cellular protocols~\cite{3gpp_36212_std, 3gpp_38212_std} utilize Polar codes for control channels with dozens of code rates, and Turbo and LDPC codes for data channels with over a hundred code rates. However, the puncturing arrangements in these standards may not always be optimal, as they are often designed for implementation convenience rather than performance maximization. Extrinsic information transfer (EXIT) charts provide a powerful semi-analytical tool for designing and analyzing iteratively decoded systems, including the optimization of puncturing schemes for Turbo codes~\cite{exit_analysis0, exit_analysis1, exit_analysis2, exit_analysis3}.

Recent studies have proposed neural network-based decoders to address puncturing in linear block codes, demonstrating improvements over traditional methods. For instance,~\cite{lbc_ecct} employed a Transformer-based network to decode linear block codes with varying code rates, while~\cite{lbc_uecct} introduced a unified Transformer decoder capable of simultaneously decoding multiple code rates of linear block codes with a single set of neural network parameters.

Other studies have focused on applying neural networks to decode sequential codes, such as convolutional and Turbo codes. For example,~\cite{comm_alg_via_dl} explored the use of recurrent neural networks (RNNs) for decoding convolutional codes, achieving performance comparable to the Viterbi algorithm~\cite{viterbi_alg_1967} in both AWGN and non-AWGN channels with $t$-distributed noise. The DeepTurbo approach~\cite{deep_turbo}, inspired by the iterative Bahl-Cocke-Jelinek-Raviv (BCJR) algorithm~\cite{bcjr_alg_1974}, targets Turbo codes but faces performance degradation when generalized to longer code lengths, requiring retraining. A notable study~\cite{mind_model_independent} employed model-agnostic meta-learning to enhance the generalization of neural network-based decoders in unseen channel conditions. Additionally, Turbo autoencoders~\cite{turbo_autoencoder} introduced an end-to-end learning framework that jointly optimizes both the encoder and decoder, outperforming traditional BCJR decoders under canonical channels.

However, these studies on convolutional and Turbo codes have not addressed the issue of puncturing, limiting their applicability in real-world communication systems. In the context of NN-based decoders, models trained without considering puncturing may fail when exposed to such conditions. This is because the model's parameters are optimized for scenarios without punctured data, making it less robust in real-world applications where puncturing is common. Furthermore, training a separate neural network for each possible code rate leads to significant storage overhead, which is impractical for scalable deployment in dynamic environments.

\subsection{Contributions}
To address this gap of puncturing issues in sequential codes, we propose a unified, protocol-compatible long short-term memory (LSTM)-based neural decoder for punctured convolutional and Turbo codes. This approach is inspired by the similarity between the LSTM’s memory mechanism and the shift register’s role in convolutional codes, as both retain a history of prior inputs to inform current outputs. The proposed approach encodes puncturing patterns directly into the neural network's latent space and introduces a balanced bit error rate training (BBT) method for efficient fine-tuning across different code rates, ensuring seamless compatibility with protocol flexibility\footnote{Simulation codes will be provided to reproduce the results in this paper: \url{http://oa.ee.tsinghua.edu.cn/dailinglong/publications/publications.html.}}. The main contributions of this work are summarized as follows.
\begin{enumerate}
\item \textbf{Protocol-Compatible Neural Decoding}:
We propose a unified protocol-compatible neural decoding approach specifically designed for punctured convolutional and Turbo codes. This approach ensures compliance with Wi-Fi (IEEE 802.11) and cellular (3GPP TS 36.212) standards by supporting varying code lengths and rates, even under practical puncturing patterns. The protocol compatibility is achieved through \textit{puncturing-aware embedding} and \textit{balanced bit error rate training}, which enables the decoder to handle different code rates effectively. To the best of our knowledge, this is the first neural decoding approach capable of generalizing seamlessly across diverse protocol-compliant configurations while maintaining competitive performance.

\item \textbf{Puncturing-Aware Embedding for Adaptability:}
We introduce a puncturing-aware embedding module that encodes puncturing patterns directly into the neural network’s latent space, enabling seamless adaptation to various code rates. By incorporating the puncturing-aware embedding as a gating mechanism, the flow of log-likelihood ratio information is controlled. This design allows the decoder to support diverse code lengths and rates with a single set of network parameters, ensuring exceptional generalization and compatibility with dynamically changing protocol requirements, making it highly suitable for modern wireless communication systems where adaptability is critical.

\item \textbf{Balanced Bit Error Rate Training}:
To further enhance flexibility and prevent overfitting to any specific code rate, we propose a balanced bit error rate training strategy. This method adjusts the signal-to-noise ratio (SNR) for different code rates to maintain a consistent bit error rate (BER) across all settings during training. As a result, each code rate contributes equally, helping the neural network generalize effectively without overfitting to any particular configuration.

\item \textbf{Superior Performance in Practical Channels}:
The proposed neural decoder demonstrates state-of-the-art performance in both additive white Gaussian noise (AWGN) and Rayleigh fading channels. Extensive simulations show that the model not only outperforms traditional decoding algorithms under practical least squares (LS) channel estimation but also exceeds the performance of conventional decoders with perfect channel state information (PCSI). Moreover, under matched code length and rate conditions during training and inference, the proposed neural decoder yields a 0.2 dB performance gain over DeepTurbo~\cite{deep_turbo} at a BER of $10^{-4}$. In mismatched conditions, it exhibits substantial performance improvements over DeepTurbo.

\end{enumerate}

This paper is structured as follows. Section II offers foundational knowledge, covering the basics of convolutional codes, Turbo codes, and long short-term memory networks. Section III describes the proposed convolutional neural engine in detail. Section IV discusses the outcomes of training and inference experiments. Section V evaluates the computational complexity and decoding latency of the proposed approach. Finally, Section VI provides a conclusion.

\textit{Notations}: Bold lowercase and uppercase letters are used to represent vectors and matrices, respectively. The symbol $\mathbb{R}$ represents the set of real numbers, while $\mathbb{C}$ denotes the set of complex numbers. The transpose operation is denoted by $[\cdot]^T$, and $[\cdot]^H$ represents the Hermitian (conjugate transpose) operation. The notation $\mathcal{N}(\mu, {\sigma}^2)$ denotes a Gaussian distribution with mean $\mu$ and variance ${\sigma}^2$.

\section{Background}
To provide a comprehensive understanding of the proposed convolutional neural engine, this section presents background knowledge on convolutional codes, Turbo codes, and long short-term memory neural networks, along with an overview of the simulation link architecture and channel model used in the system.

\subsection{Fundamentals of Convolutional and Turbo Codes}

\subsubsection{Convolutional Codes in IEEE 802.11}
\
\newline
\indent
IEEE 802.11 mandates convolutional coding as a required feature for both access points (APs) and stations (STAs) in the physical layer (PHY), specifically within the orthogonal frequency division multiplexing (OFDM) modulation schemes. The standard supports convolutional codes with rates of 1/2, 2/3, 3/4, and 5/6 for forward error correction (FEC) to mitigate bit errors caused by noise, fading, and interference.
\begin{equation}
\label{eq_bcc_gen_poly}
\begin{aligned}
G = \left[ {{g_0} = {{133}_8},{g_1} = {{171}_8}} \right]
\end{aligned}
\end{equation}

\begin{figure}[htbp]
\centering
\includegraphics[width=0.5\columnwidth]{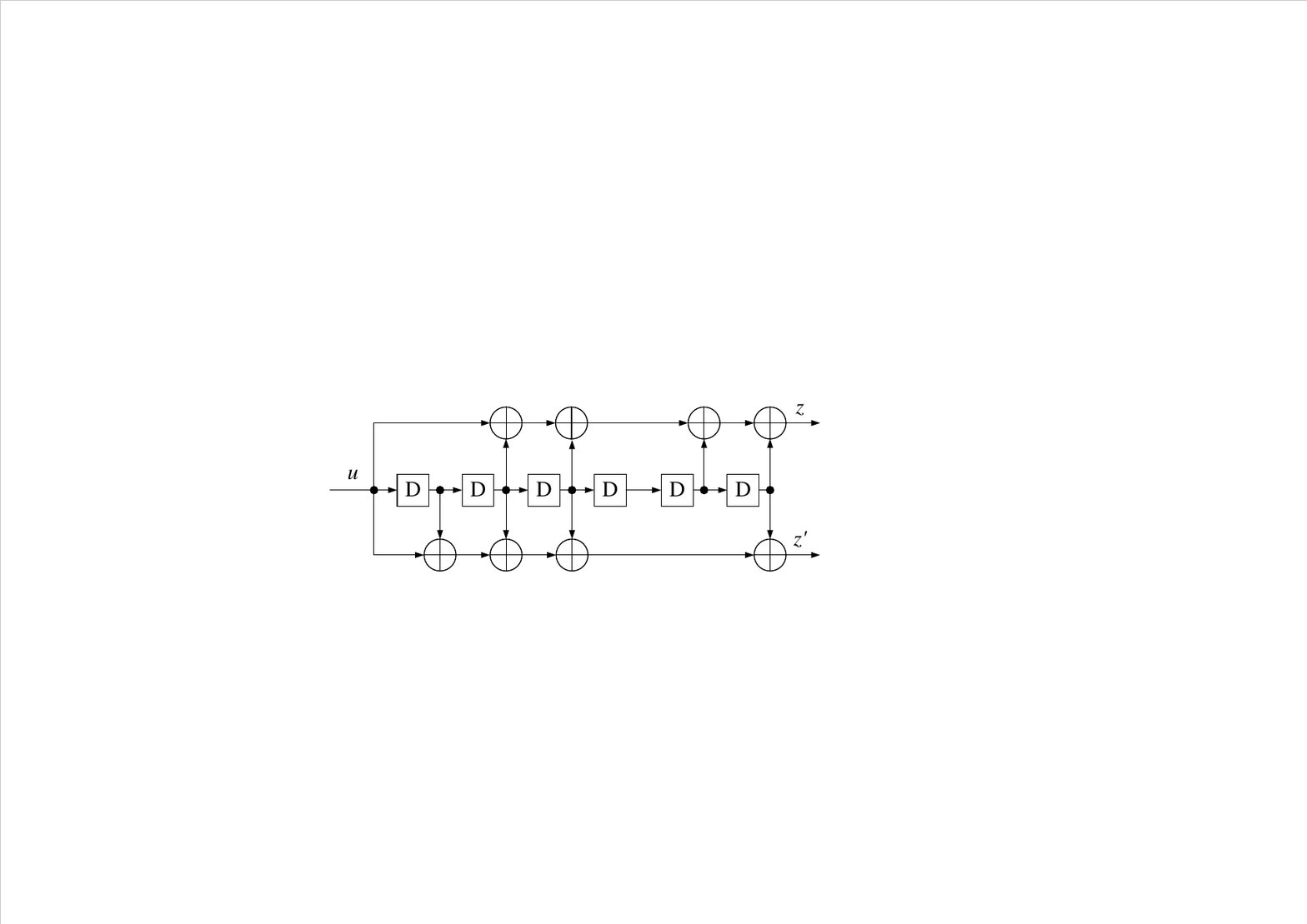}
\caption{Encoding structure of the convolutional code in IEEE 802.11.}
\label{fig_bcc_encoder}
\end{figure}

\begin{figure}[htbp]
\centering
\includegraphics[width=0.5\columnwidth]{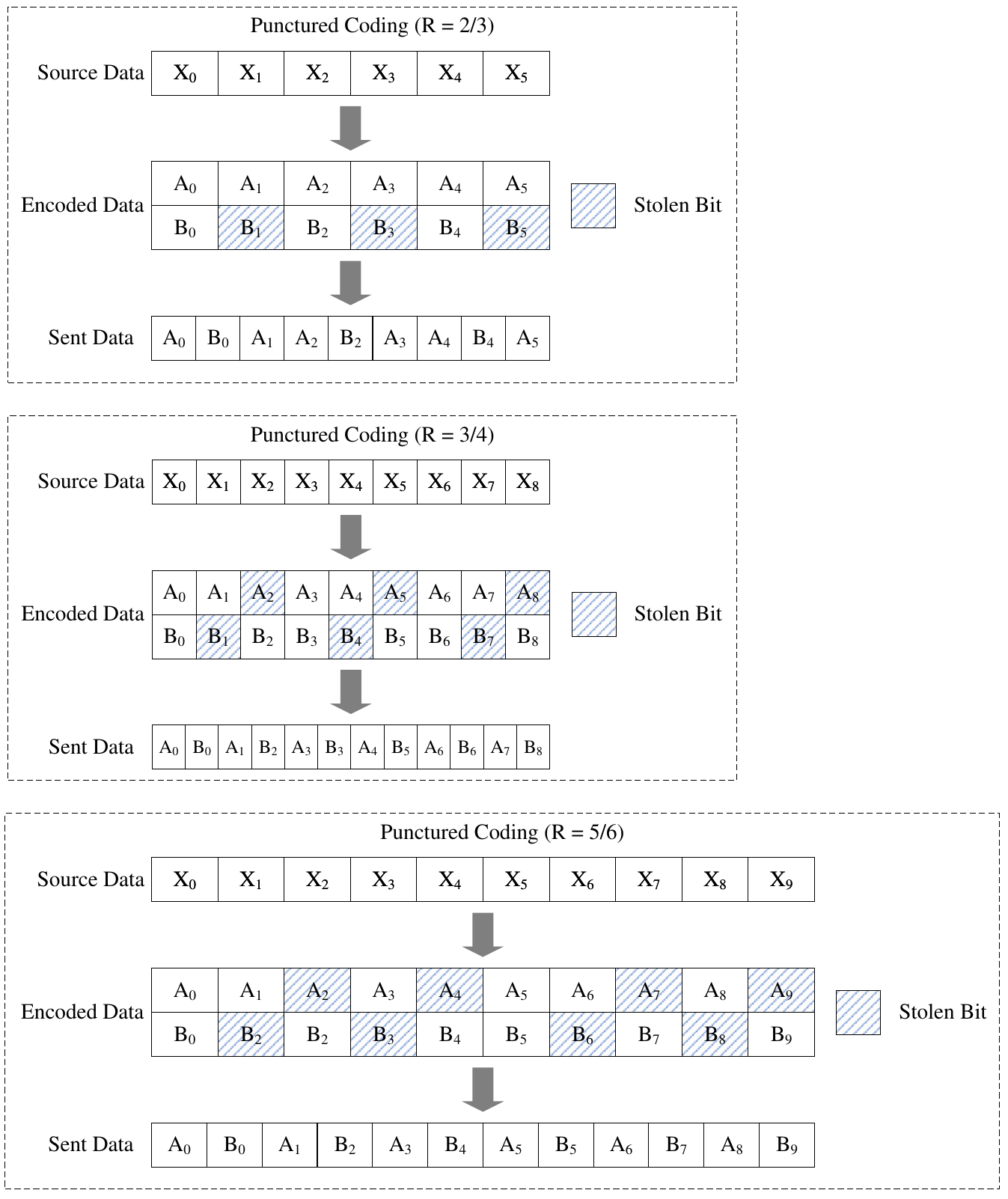}
\caption{Example of the bit-stealing procedure in IEEE 802.11 (R = 2/3, 3/4, 5/6).}
\label{fig_bcc_puncture}
\end{figure}

The convolutional code is characterized by its encoder structure, where the input bit stream passes through a series of shift registers, generating encoded bits as output. Each output bit is determined by the current input bit and several previous input bits, as defined by the constraint length of the code. The constraint length represents the number of bits stored in the encoder's memory that influence the current output. In IEEE 802.11, the convolutional encoder operates with a mother code rate of 1/2 and a constraint length of 7. This setup produces two output bits for each input bit, with the constraint length indicating that the output depends on the current bit and the six preceding bits. Figure~\ref{fig_bcc_encoder} provides a schematic illustration of the convolutional encoder used in IEEE 802.11, showcasing its encoding process and structure. The bit denoted as ``$z$'' shall be output from the encoder before the bit denoted as ``$z'$''. The specific generator polynomials governing this encoding process, expressed in octal form, are provided in Eq.~\ref{eq_bcc_gen_poly}. Higher code rates, such as 2/3, 3/4, and 5/6, are achieved using a technique known as puncturing. Puncturing selectively removes specific encoded bits to increase the effective code rate without altering the encoder's fundamental structure. Figure~\ref{fig_bcc_puncture} illustrates how puncturing is applied in IEEE 802.11 to achieve these higher code rates.

\subsubsection{Turbo Codes in 3GPP TS 36.212}
\
\newline
\indent
Turbo codes, introduced in the late 1990s, represent a significant advancement in error correction techniques. These codes are used in the 3rd generation partnership project (3GPP) standards, specifically in the 3GPP TS 36.212 specification for long-term evolution (LTE) and beyond.
\begin{equation}
\label{eq_turbo_gen_poly}
\begin{aligned}
G\left( D \right) = \left[ {1,\frac{{{g_1}\left( D \right)}}{{{g_0}\left( D \right)}}} \right]{,_{}}\left\{ {\begin{array}{*{20}{l}}
{{g_0}\left( D \right) = 1 + {D^2} + {D^3}}\\
{{g_1}\left( D \right) = 1 + D + {D^3}}
\end{array}} \right.
\end{aligned}
\end{equation}
\begin{figure}[htbp]
\centering
\includegraphics[width=0.5\columnwidth]{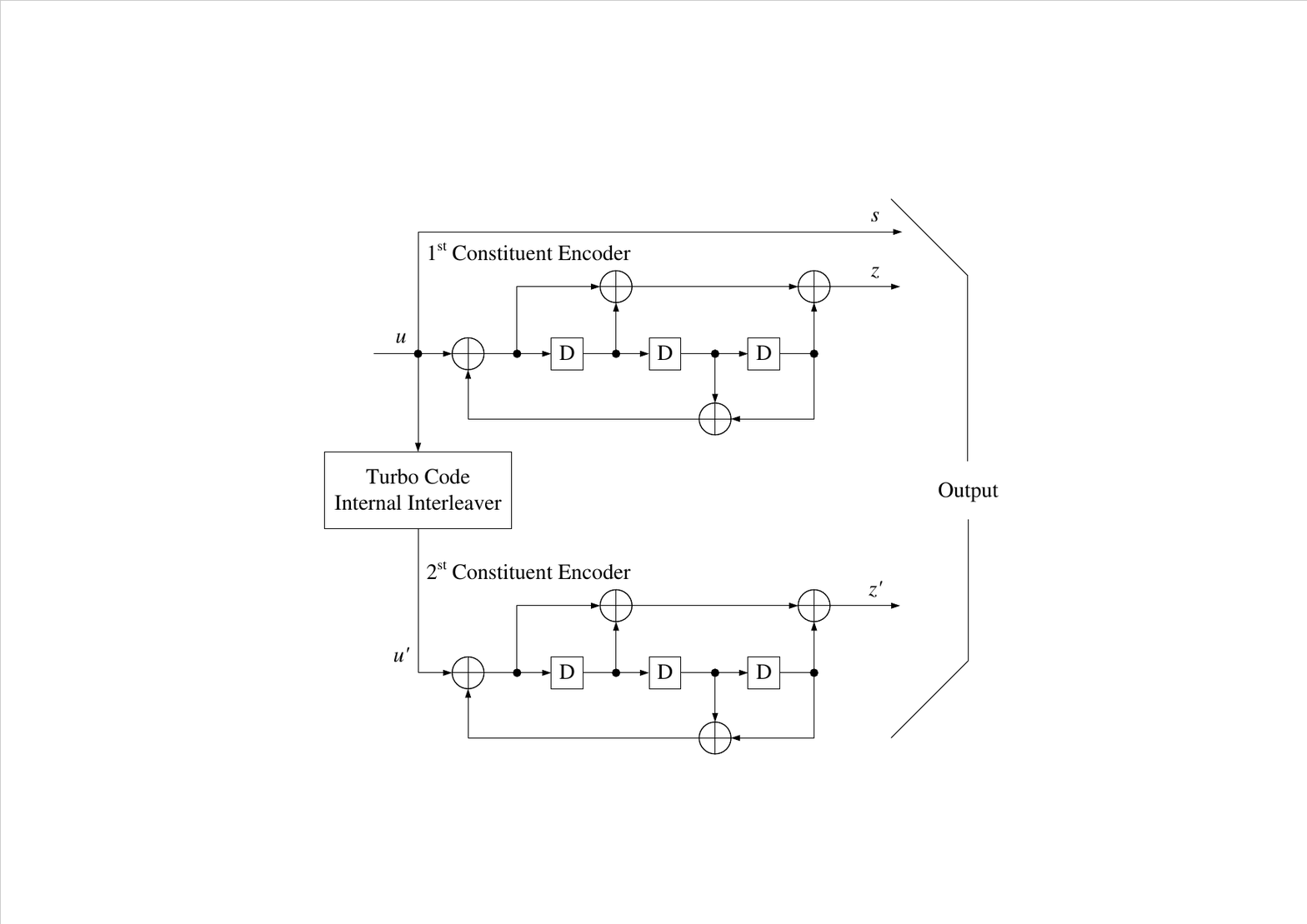}
\caption{Encoding structure of the Turbo codes in 3GPP TS 36.212.}
\label{fig_turbo_encoder}
\end{figure}
\begin{figure}[htbp]
\centering
\includegraphics[width=0.5\columnwidth]{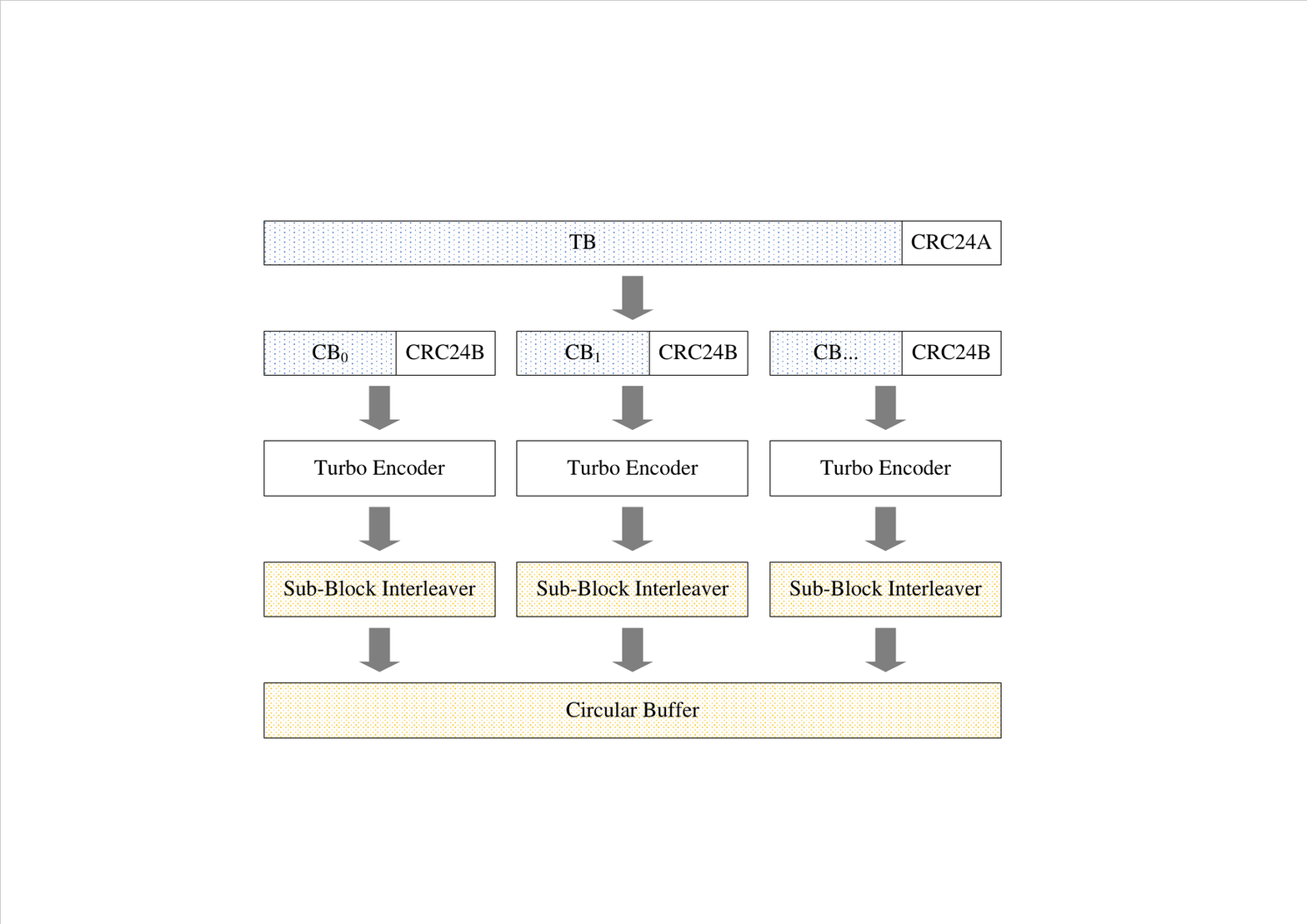}
\caption{Example of the rate matching procedure in 3GPP TS 36.212.}
\label{fig_turbo_rate_matching}
\end{figure}

In 3GPP TS 36.212, Turbo codes are used in the downlink and uplink channels for the physical layer, where they provide strong error correction capabilities. The standard adopts parallel concatenated convolutional codes (PCCC) as the foundation of Turbo coding. Figure~\ref{fig_turbo_encoder} illustrates the encoding process of Turbo codes, which consists of two parallel convolutional encoders, each with a constraint length of 4. The first encoder processes the input bitstream directly, while the second operates on an interleaved version of the input, with the interleaver introducing randomness to the input sequence. The specific generator polynomials for these encoders are detailed in Equation~\ref{eq_turbo_gen_poly}.

The mother code rate of Turbo codes is 1/3, producing three output bits for each input bit. To support higher data rates, the effective code rate can be increased up to 0.93 through puncturing, which selectively omits specific encoded bits. This enables flexibility in adapting to different data rate requirements while maintaining error correction capabilities. The intermediate code rates between the mother code rate and the maximum rate are achieved by adjusting the allocation of time-frequency physical resources based on channel quality indicators (CQIs) reported by the user. This adaptive process, known as rate matching, enables efficient code rate adjustment to ensure optimal performance under varying channel conditions.

Figure~\ref{fig_turbo_rate_matching} provides a comprehensive overview of the rate matching process, which adapts the encoded data to the available physical resources. This process begins with the transport block (TB) being divided into code blocks (CBs) after appending a 24-bit cyclic redundancy check (CRC) using CRC24A for error detection. Each CB then adds another 24-bit CRC24B and is subsequently processed by the Turbo encoder. The encoded CBs undergo sub-block interleaving, which rearranges the data to improve resilience against burst errors. The interleaved code blocks are then processed through a circular buffer, which facilitates the rate matching procedure by selecting and concatenating bits as per the allocated resource block size. This process is crucial for aligning the data to the physical layer's constraints while ensuring efficient utilization of resources.

\subsection{Long Short-Term Memory Neural Network}
\begin{figure}[htbp]
\centering
\includegraphics[width=0.5\columnwidth]{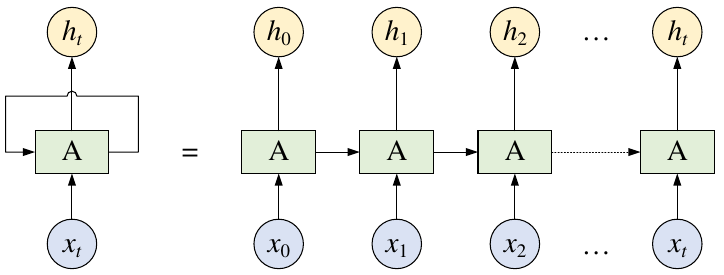}
\caption{The unrolled recurrent neural network.}
\label{fig_rnn_structure}
\end{figure}

Long short-term memory networks~\cite{lstm_nn}, a variant of recurrent neural networks~\cite{rnn_network}, are specifically designed to model sequential data and capture long-term dependencies effectively. As shown in Figure~\ref{fig_rnn_structure}, RNNs employ a cyclic structure that allows information to persist across time steps $t$, enabling sequential data processing. The update of the hidden state $\boldsymbol{h}_t$ in an RNN is governed by:
\begin{equation}
\label{eq_rnn}
\begin{aligned}
\boldsymbol{h}_t = \tanh \left( \boldsymbol{W}_h \boldsymbol{h}_{t-1} + \boldsymbol{W}_x \boldsymbol{x}_t + \boldsymbol{b}_h \right)
\end{aligned}
\end{equation}
where $\boldsymbol{h}_t$ is the hidden state at time step $t$, $\boldsymbol{h}_{t-1}$ is the hidden state from the previous time step, $\boldsymbol{x}_t$ is the input at time step $t$, and $\boldsymbol{W}_h$, $\boldsymbol{W}_x$, $\boldsymbol{b}_h$ are learnable parameters. This cyclic nature of RNNs makes them well-suited for capturing temporal dependencies in sequential data. However, traditional RNNs encounter challenges during training, particularly vanishing and exploding gradient issues, which hinder their ability to capture long-range dependencies.
\begin{figure}[htbp]
\centering
\includegraphics[width=0.5\columnwidth]{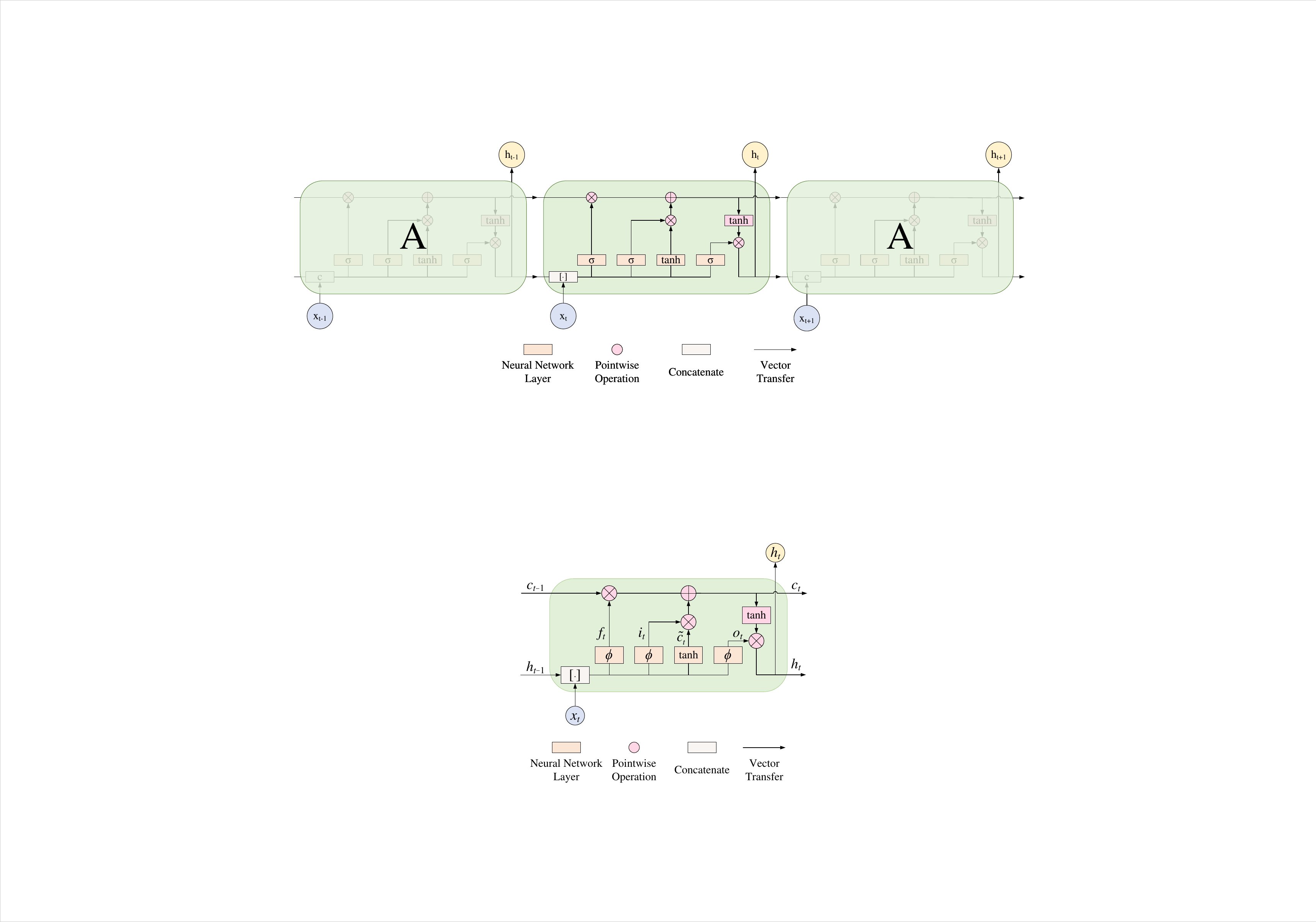}
\caption{The architecture of an LSTM cell.}
\label{fig_lstm_structure}
\end{figure}

LSTMs, introduced by Hochreiter and Schmidhuber in 1997~\cite{lstm_nn}, overcome these limitations through an innovative memory cell structure. Unlike the simpler architecture of RNNs, LSTMs incorporate a gating mechanism to dynamically regulate information flow, enabling the network to selectively retain, update, or discard information. The architecture of an LSTM cell, illustrated in Figure~\ref{fig_lstm_structure}, includes three core gates: the forget gate, input gate, and output gate. These gates are mathematically defined as follows:
\begin{equation}
\label{eq_lstm}
\begin{aligned}
\begin{array}{l}
\boldsymbol{f}_t = \phi \left( \boldsymbol{W}_f \cdot \begin{bmatrix} \boldsymbol{h}_{t-1} , \boldsymbol{x}_t \end{bmatrix} + \boldsymbol{b}_f \right) \\

\boldsymbol{i}_t = \phi \left( \boldsymbol{W}_i \cdot \begin{bmatrix} \boldsymbol{h}_{t-1} , \boldsymbol{x}_t \end{bmatrix} + \boldsymbol{b}_i \right) \\

\boldsymbol{o}_t = \phi \left( \boldsymbol{W}_o \cdot \begin{bmatrix} \boldsymbol{h}_{t-1} , \boldsymbol{x}_t \end{bmatrix} + \boldsymbol{b}_o \right) \\

\tilde{\boldsymbol{c}}_t = \tanh \left( \boldsymbol{W}_c \cdot \begin{bmatrix} \boldsymbol{h}_{t-1} , \boldsymbol{x}_t \end{bmatrix} + \boldsymbol{b}_c \right) \\

\boldsymbol{c}_t = \boldsymbol{f}_t \odot \boldsymbol{c}_{t-1} + \boldsymbol{i}_t \odot \tilde{\boldsymbol{c}}_t \\

\boldsymbol{h}_t = \boldsymbol{o}_t \odot \tanh \left( \boldsymbol{c}_t \right)
\end{array}
\end{aligned}
\end{equation}
where $\boldsymbol{f}_t$, $\boldsymbol{i}_t$, and $\boldsymbol{o}_t$ represent the forget, input, and output gate activations at time step $t$; $\boldsymbol{x}_t$ is the input;  $\tilde{\boldsymbol{c}}_t$ is the candidate memory cell; $\boldsymbol{c}_t$ is the cell state; and $\boldsymbol{h}_t$ is the hidden state.

Here, $\phi$ denotes the sigmoid activation function, which plays a critical gating role by producing values in the range of $[0, 1]$. This enables $\phi$ to act as a ``soft switch'' that determines the degree to which information should be passed through the network. For instance, in the forget gate $\boldsymbol{f}_t$, $\phi$ dynamically adjusts the proportion of the previous cell state $\boldsymbol{c}_{t-1}$ that should be retained. Similarly, in the input gate $\boldsymbol{i}_t$ and output gate $\boldsymbol{o}_t$, $\phi$ controls how much new information is incorporated into the cell state and how much of the cell state influences the hidden state $\boldsymbol{h}_t$, respectively. By gating these information flows, $\phi$ ensures that the network can focus on relevant inputs and maintain stable gradients during training, allowing LSTMs to effectively capture both short- and long-term dependencies.

\subsection{System Model}
The proposed approach is evaluated under two distinct communication channel models: the AWGN channel and the Rayleigh fading channel. The AWGN channel serves as the training and inference environment, providing an idealized setting for model training. In contrast, the Rayleigh fading channel is used solely for inference, allowing us to assess how well the model, trained in the AWGN channel, generalizes to more realistic fading conditions.

\begin{figure*}[htbp]
\centering
\includegraphics[width=1.0\textwidth]{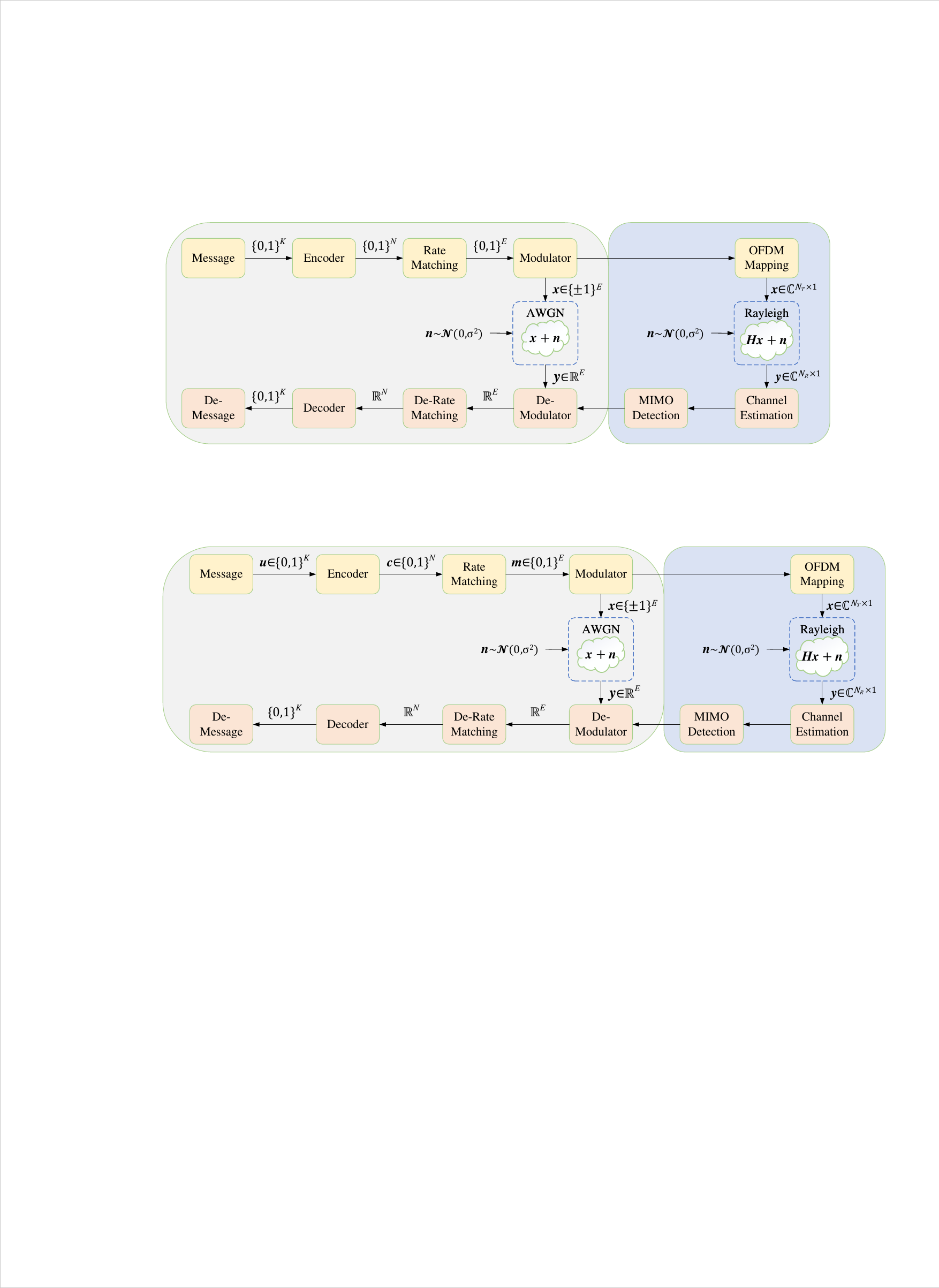}
\caption{The communication link of the AWGN and Rayleigh fading channel.}
\label{fig_awgn_rayleigh_channel_model}
\end{figure*}

In the AWGN channel, the communication link follows the sequence illustrated in Figure~\ref{fig_awgn_rayleigh_channel_model}, where $K$ denotes the length of the original message, $N$ represents the length of the encoded message, and $E$ refers to the length of the message after rate-matching. The code rate $R$ is defined as the ratio of the original message length to the rate-matched length, i.e., $R = \frac{K}{E}$.

The rate-matched message $\boldsymbol{m} \in \{0,1\}^E$ is modulated using binary phase shift keying (BPSK) to produce the transmitted signal $\boldsymbol{x} \in \{\pm 1\}^E$. The received signal $\boldsymbol{y} \in \mathbb{R}^E$ is corrupted by additive Gaussian noise $\boldsymbol{n} \in \mathbb{R}^E$, and the relationship between the transmitted and received signals is expressed as:
\begin{equation}
\label{eq_awgn_model}
\begin{aligned}
\boldsymbol{y} = \boldsymbol{x} + \boldsymbol{n}
\end{aligned}
\end{equation}
where $\boldsymbol{n} \sim \mathcal{N}(0, \sigma^2)$ represents the AWGN noise. After demodulation and de-rate matching, the signal is decoded to obtain the original message $\boldsymbol{u} \in \{0,1\}^K$.

In contrast, the Rayleigh fading channel is utilized for inference to simulate more realistic wireless communication conditions, where the transmitted signal experiences multipath fading. The communication link for the Rayleigh fading channel is detailed in Figure~\ref{fig_awgn_rayleigh_channel_model}, which illustrates the multipath fading model and frequency-domain processing.

The received signal in the Rayleigh fading channel is modeled as:
\begin{equation}
\label{eq_rayleigh_model}
\begin{aligned}
\boldsymbol{y} = \boldsymbol{H} \boldsymbol{x} + \boldsymbol{n}
\end{aligned}
\end{equation}
where $\boldsymbol{H} \in \mathbb{C}^{N_R \times N_T}$ is the channel matrix representing the Rayleigh fading coefficients, with $N_R$ denoting the number of receive antennas and $N_T$ the number of transmit antennas. $\boldsymbol{x} \in \mathbb{C}^{N_T \times 1}$ is the transmitted signal vector in the frequency domain after modulation (e.g., OFDM subcarriers). $\boldsymbol{n} \in \mathbb{C}^{N_R \times 1}$ is the noise vector at the receiver, where each element is i.i.d. Gaussian noise with zero mean and variance $\sigma^2$.

The channel matrix $\boldsymbol{H}$ is constructed by generating multipath coefficients in the time domain, based on an $L$-tap delay profile determined by environmental characteristics and delay spread, and then transformed into the frequency domain via the fast fourier transform (FFT) to be applied to $\boldsymbol{x}$. Each tap of $\boldsymbol{H}$ is modeled as an independent complex Gaussian random variable, and the total power of the $L$ taps is normalized to 1, i.e., $\sum_{l=0}^{L-1} \mathbb{E}[|h_l|^2] = 1$, where $h_l$ represents the coefficient of the $l$-th tap.

Channel estimation is performed to estimate $\boldsymbol{H}$, and multiple input multiple output (MIMO) detection is used to obtain the estimated transmitted signal $\hat{\boldsymbol{x}}$, after $\boldsymbol{x}$ has been affected by multipath fading. The subsequent steps, including demodulation, de-rate matching, and decoding, are similar to those in the AWGN case, with the goal of recovering the original message.

By training the model on the AWGN channel and testing it on the Rayleigh fading channel, we evaluate how well the decoder generalizes to real-world, challenging environments.

\section{Proposed Convolutional Neural Engine}
In this section, we present the complete architecture and training process of the proposed \textbf{convolutional neural engine (CNE)}.

\subsection{Overall Architecture}
\begin{figure*}[htbp]
\centering
\includegraphics[width=1.0\textwidth]{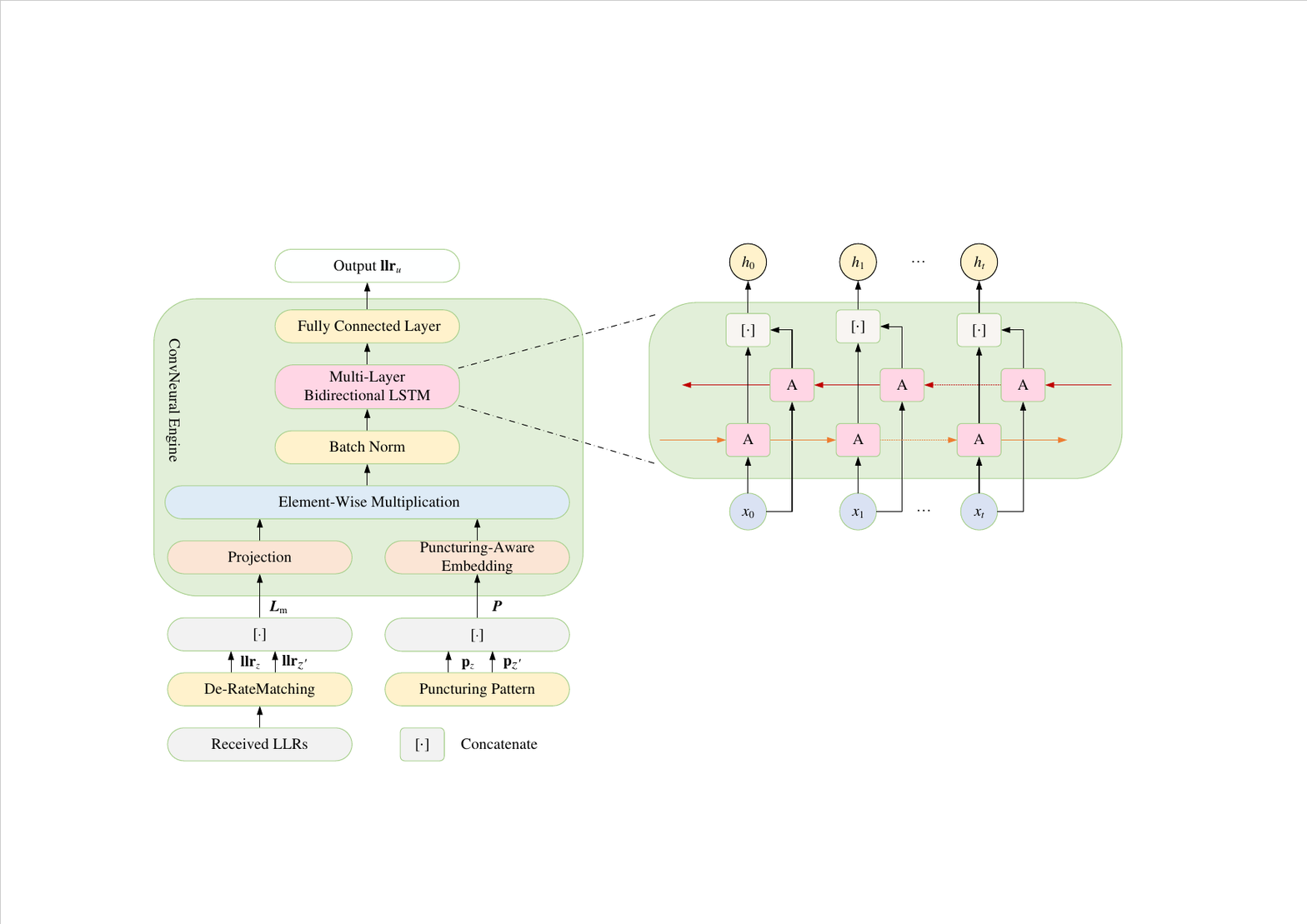}
\caption{The architecture of the proposed convolutional neural engine.}
\label{fig_bcc_decoder_structure}
\end{figure*}

The proposed convolutional neural engine introduces a groundbreaking approach to decoding convolutional codes, particularly under punctured conditions, by effectively integrating domain-specific insights into a deep learning framework. Unlike conventional NN-based methods, which often treat punctured sequences as missing data, CNE incorporates an innovative \textbf{puncturing-aware embedding} mechanism that explicitly encodes puncturing patterns into the feature space. This approach enhances the decoder's adaptability, enabling it to generalize across varying code rates specified by protocols and ensuring compatibility for practical applications. The architecture of the proposed CNE is illustrated in Figure~\ref{fig_bcc_decoder_structure}.

\subsubsection{Convolutional Decoding}
\
\newline
\indent
Before decoding, the received log-likelihood ratios (LLRs) are rate-dematched by inserting zeros at the positions of punctured bits to restore the original codeword length $N$. The vectors $\mathbf{llr}_{z} \in \mathbb{R}^{K}$ and $\mathbf{llr}_{z'} \in \mathbb{R}^{K}$, together with their corresponding puncturing indicators $\mathbf{p}_{z} \in \{0,1\}^{K}$ and $\mathbf{p}_{z'} \in \{0,1\}^{K}$, are concatenated to form the matrix $\boldsymbol{L}_{\text{m}} \in \mathbb{R}^{K \times 2}$ and its associated puncturing pattern $\boldsymbol{P} \in \{0,1\}^{K \times 2}$. These matrices serve as the input to the convolutional neural engine.

At the core of the CNE architecture lies a sequence of operations that transform raw input data and puncturing patterns into actionable decoding outputs. The process begins with a projection of the input sequence $\boldsymbol{L}_{\text{m}}$ into a higher-dimensional embedding space:
\begin{equation}
\label{eq_lstm_input_embedding}
\begin{aligned}
\boldsymbol{E}_{l} = \boldsymbol{W}_{\text{proj}} \boldsymbol{L_{\text{m}}} + \boldsymbol{b}_{\text{proj}}
\end{aligned}
\end{equation}
where $\boldsymbol{W}_{\text{proj}} \in \mathbb{R}^{D_{\text{embed}} \times D_{\text{in}}}$ and $\boldsymbol{b}_{\text{proj}} \in \mathbb{R}^{D_{\text{embed}}}$ are learnable parameters, $K$ represents the sequence length, and the input dimension $D_{\text{in}} = 2$ corresponds to the LLRs of the two coded bits generated per time step in IEEE 802.11 convolutional codes.

A \textbf{puncturing-aware embedding} module integrates puncturing information by mapping the pattern $\boldsymbol{P} \in \{0,1\}^{K \times D_{\text{in}}}$, where each element specifies whether the corresponding position in $\boldsymbol{L}_{\text{m}}$ is punctured (0) or non-punctured (1), into the embedding space of $\boldsymbol{E}_{p}$. This is achieved using:
\begin{equation}
\label{eq_lstm_puncture_embedding_0}
\begin{aligned}
\boldsymbol{E}_{p} = \phi \left( \boldsymbol{W}_{\text{punc}} \boldsymbol{P} + \boldsymbol{b}_{\text{punc}} \right)
\end{aligned}
\end{equation}
where $\boldsymbol{W}_{\text{punc}} \in \mathbb{R}^{D_{\text{embed}} \times D_{\text{in}}}$, $\boldsymbol{b}_{\text{punc}} \in \mathbb{R}^{D_{\text{embed}}}$, and $\phi(\cdot)$ is the sigmoid activation function, which ensures that the puncture information acts as a gate, controlling the flow of puncturing effects.

The input embedding $\boldsymbol{E}_{l}$ and puncturing-aware embedding $\boldsymbol{E}_{p}$ are then combined through element-wise multiplication:
\begin{equation}
\label{eq_lstm_puncture_embedding_1}
\begin{aligned}
\boldsymbol{E}_{lp} = \boldsymbol{E}_{l} \odot \boldsymbol{E}_{p}
\end{aligned}
\end{equation}
allowing the model to dynamically adjust the input representation based on the puncturing pattern.

Following this, the combined embedding $\boldsymbol{E}_{lp}$ undergoes batch normalization (BN) to stabilize training:
\begin{equation}
\label{eq_lstm_bn}
\begin{aligned}
\boldsymbol{E}_{\text{norm}} = \text{BN}(\boldsymbol{E}_{lp})
\end{aligned}
\end{equation}

The normalized embedding is then processed by a multi-layer bidirectional LSTM, which captures both forward and backward temporal dependencies in the sequence:
\begin{equation}
\label{eq_lstm_core}
\begin{aligned}
\boldsymbol{S}_{\text{out}} = \text{LSTM}(\boldsymbol{E}_{\text{norm}})
\end{aligned}
\end{equation}
where $\boldsymbol{S}_{\text{out}} \in \mathbb{R}^{K \times 2D_{\text{hidden}}}$ represents the encoded context of the entire sequence, leveraging information from both directions.

Finally, the output from the LSTM is passed through a fully connected layer to produce the decoded bit likelihoods:
\begin{equation}
\label{eq_lstm_output}
\begin{aligned}
\mathbf{llr}_{u} = \boldsymbol{W}_{\text{out}} \boldsymbol{S}_{\text{out}} + \boldsymbol{b}_{\text{out}}
\end{aligned}
\end{equation}
where $\boldsymbol{W}_{\text{out}} \in \mathbb{R}^{2D_{\text{hidden}} \times 1}$, and $\mathbf{llr}_{u} \in \mathbb{R}^{K}$ represents the likelihoods of the estimated bits being 1.

The entire decoding process can be summarized as:
\begin{equation}
\label{eq_lstm_total_process}
\begin{aligned}
\mathbf{llr}_{u} =  \; \boldsymbol{W}_{\text{out}} \cdot \text{LSTM} \Big( \text{BN} \Big( \big( \boldsymbol{W}_{\text{proj}} \boldsymbol{L}_{\text{m}} + \boldsymbol{b}_{\text{proj}} \big) \odot 
 \phi \big( \boldsymbol{W}_{\text{punc}} \boldsymbol{P} + \boldsymbol{b}_{\text{punc}} \big) \Big) \Big) + \boldsymbol{b}_{\text{out}}
\end{aligned}
\end{equation}

\subsubsection{Turbo Decoding}
\begin{figure}[htbp]
\centering
\includegraphics[width=0.5\columnwidth]{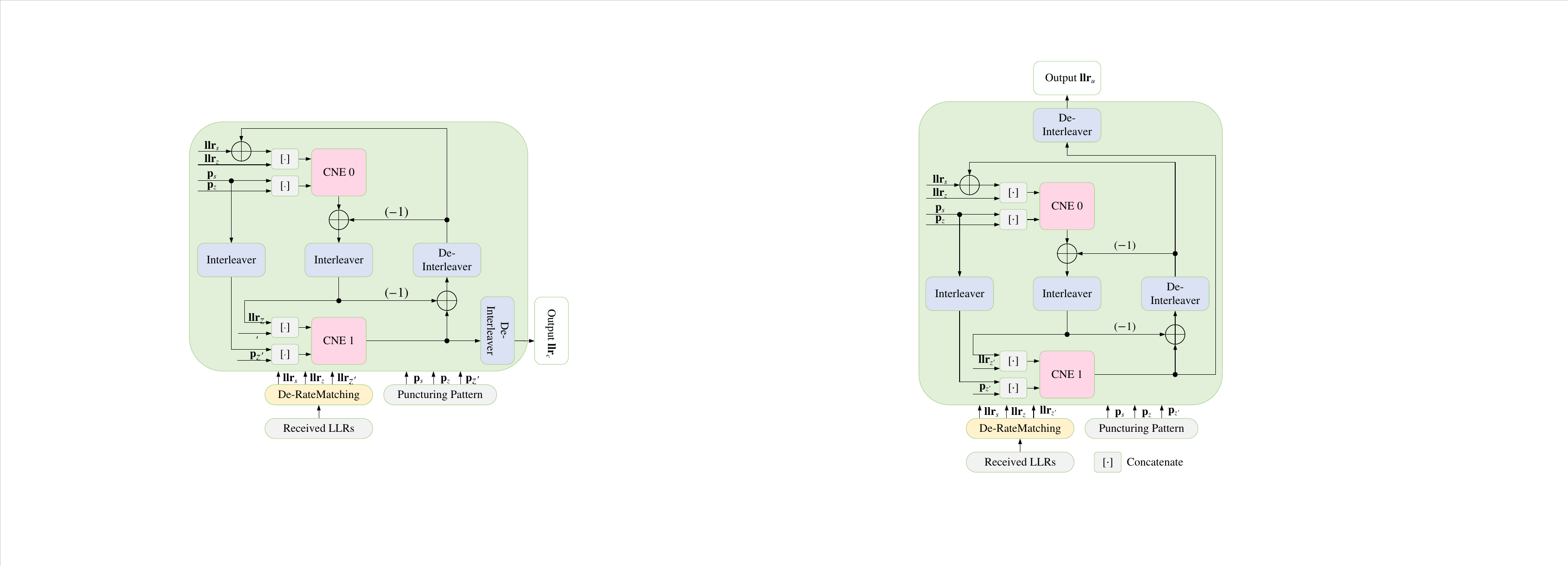}
\caption{The architecture of the LSTM-based Turbo code decoder.}
\label{fig_turbo_decoder_structure}
\end{figure}

\
\newline
\indent
The proposed LSTM-based Turbo decoder builds upon the principles of traditional BCJR Turbo decoding, employing two identical CNEs as its core components. These CNEs are connected through interleaving and de-interleaving mechanisms, enabling iterative refinement of the decoding process. The architecture is specifically designed to capture both the sequential nature of convolutional codes and the iterative exchange of extrinsic information that characterizes Turbo decoding. This architecture is also visualized in Figure~\ref{fig_turbo_decoder_structure}, which illustrates the interaction between the two component CNEs.

Before the iterative decoding process begins, the received sequence undergoes a rate-matching reversal procedure, which consists of de-puncturing and sub-block de-interleaving operations. During de-puncturing, zeros are inserted at the locations of punctured bits to restore the original codeword structure. These processes generate the systematic sequence $\mathbf{llr}_{s} \in \mathbb{R}^{K}$, the first and second parity sequences $\mathbf{llr}_{z} \in \mathbb{R}^{K}$ and $\mathbf{llr}_{z'} \in \mathbb{R}^{K}$, along with their corresponding puncturing indicators $\mathbf{p}_{s} \in \{0,1\}^{K}$, $\mathbf{p}_{z} \in \{0,1\}^{K}$, and $\mathbf{p}_{z'} \in \{0,1\}^{K}$, which serve as inputs to the CNE-based Turbo code decoder.

In our designed architecture \textbf{CNE 0} processes the systematic sequence $\mathbf{llr}_{s}$ alongside the first parity sequence $\mathbf{llr}_{z}$. \textbf{CNE 1} processes the interleaved systematic bit sequence and the second parity sequence $\mathbf{llr}_{z'}$. The two CNEs share identical architectures and weights, ensuring symmetry and simplifying training.

The decoding process operates iteratively, exchanging extrinsic information between the two CNEs through interleaving and de-interleaving. At each iteration, the following steps occur:
\begin{equation}
\label{eq_turbo_total_process}
\begin{aligned}
&\text{Step 1.}\quad
\widehat{\mathbf{llr}}_0^{(t)} = f_{\text{CNE 0}} 
\left(
\begin{bmatrix}
\mathbf{llr}_{s} + \widehat{\mathbf{llr}}_1^{\text{de-int}, (t-1)}, \mathbf{llr}_{z}
\end{bmatrix},
\begin{bmatrix}
\mathbf{p}_{s},
\mathbf{p}_{z}
\end{bmatrix}
\right)
\\
&\text{Step 2.}\quad
\widehat{\mathbf{llr}}_0^{\text{int}, (t)} = \boldsymbol{\pi}
\left(
\widehat{\mathbf{llr}}_0^{(t)}-\widehat{\mathbf{llr}}_1^{\text{de-int}, (t-1)}
\right)
\\
&\text{Step 3.}\quad
\widehat{\mathbf{llr}}_1^{(t)} = f_{\text{CNE 1}}
\left(
\begin{bmatrix}
\widehat{\mathbf{llr}}_0^{\text{int}, (t)}, \mathbf{llr}_{z'}
\end{bmatrix},
\begin{bmatrix}
\boldsymbol{\pi}(\mathbf{p}_{s}), \mathbf{p}_{z'}
\end{bmatrix}
\right)
\\
&\text{Step 4.}\quad
\widehat{\mathbf{llr}}_1^{\text{de-int}, (t)} = \boldsymbol{\pi}^{-1}
\left(
\widehat{\mathbf{llr}}_1^{(t)}-\widehat{\mathbf{llr}}_0^{\text{int}, (t)}
\right)
\end{aligned}
\end{equation}

Here, $t$ denotes the iteration step, while $\boldsymbol{\pi}$ and $\boldsymbol{\pi}^{-1}$ represent the interleaving and de-interleaving operations, respectively. After $N_{\text{iter}}$ iterations, the final estimate of the systematic bit likelihood is obtained by applying the de-interleaving operation to the output of CNE 1:
\begin{equation}
\label{eq_turbo_final_output}
\begin{aligned}
\mathbf{llr}_{u} = \boldsymbol{\pi}^{-1} \left( \widehat{\mathbf{llr}}_1^{(N_{\text{iter}})} \right)
\end{aligned}
\end{equation}

\subsubsection{Structural Differences Between CNE and DeepTurbo}
\
\newline
\indent
\begin{figure}[htbp]
\centering
\includegraphics[width=0.5\columnwidth]{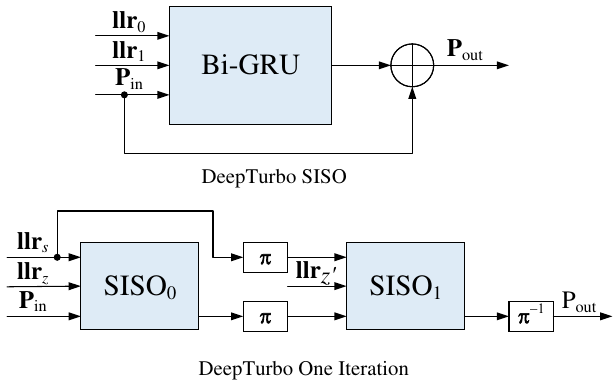}
\caption{The architecture of the DeepTurbo decoder.}
\label{fig_deep_turbo_structure}
\end{figure}
Figure~\ref{fig_deep_turbo_structure} illustrates the architecture of the DeepTurbo decoder~\cite{deep_turbo} with a bidirectional gated recurrent unit (Bi-GRU) decoding core and two soft-in soft-out (SISO) modules, where $\mathbf{P}_\text{in}$ and $\mathbf{P}_\text{out}$ represent posterior information with a feature size of $F_s$. Although both the proposed CNE and DeepTurbo employ iterative decoding inspired by the BCJR algorithm, their architectures differ significantly in three key aspects, as outlined below:
	\begin{enumerate}
	\item \textbf{Shared Weights}: Unlike DeepTurbo, where the neural network weights for the two SISO modules are independent, our proposed CNE0 and CNE1 share the same set of weights. This design reduces the total number of parameters, thereby lowering storage requirements and improving computational efficiency, making the architecture more practical for resource-constrained environments.
	\item \textbf{Enhanced High-Dimensional Embedding}: While DeepTurbo expands only the posterior information into a higher-dimensional space, our CNE-based Turbo decoder extends the systematic bits ($\mathbf{llr}_s$), as well as both parity sequences ($\mathbf{llr}_z$ and $\mathbf{llr}_{z'}$), into a high-dimensional embedding space. This comprehensive embedding approach enhances the expressive power of the neural network, enabling it to better capture complex relationships within the codeword.
	\item \textbf{Puncturing-Aware Embedding Layer}: A critical distinction of our work is the introduction of a novel puncturing-aware embedding layer, which DeepTurbo lacks. This layer explicitly incorporates puncturing patterns into the neural network’s latent space, enabling seamless generalization across various code rates. This feature ensures compatibility with practical communication protocols, such as IEEE 802.11 and 3GPP TS 36.212, where puncturing is prevalent.
	\end{enumerate}
The proposed CNE architecture overcomes DeepTurbo’s limitations in generalizing to diverse code lengths and rates, making it more suitable for modern communication requirements.

\subsection{Training Methodology}
To optimize the proposed CNE-based convolutional and Turbo decoder for diverse code rates and puncturing patterns, a two-stage training strategy is employed. The process begins with pre-training on non-punctured codes at a fixed SNR of 0 dB. This pre-training step establishes a strong foundational model capable of decoding convolutional and Turbo codes under idealized conditions, serving as the basis for subsequent fine-tuning.

Building upon this pretrained model, fine-tuning is performed using a mixed-rate dataset containing samples from multiple code rates, each associated with distinct puncturing patterns. During fine-tuning, the SNR for codewords at different code rates $R$ is adjusted using the formula:
\begin{equation}
\label{eq_mix_train_snr}
\begin{aligned}
\text{SNR}_{\text{train}} = \text{SNR}_{\text{offset}} + 10 \log_{10}(2R)
\end{aligned}
\end{equation}
where $\text{SNR}_{\text{offset}}$ serves as a baseline SNR value calibrated to align the bit error rate across code rates. This parameter can be selected based on empirical results from conventional decoders, such as the Viterbi or BCJR algorithms, to ensure approximate BER alignment across code rates. This \textbf{balanced BER training (BBT)} prevents the loss function from being dominated by specific code rates, maintaining stability during optimization.

The training loss is defined using binary cross-entropy (BCE), which evaluates the accuracy of the predicted likelihoods against the true bit labels. Mathematically, the BCE loss is expressed as:
\begin{equation}
\label{eq_train_loss}
\begin{aligned}
\mathcal{L}_{\text{BCE}} = -\frac{1}{K} \sum_{k=1}^K \left( y_k \log(\hat{y}_k) + (1 - y_k) \log(1 - \hat{y}_k) \right)
\end{aligned}
\end{equation}
where $K$ is the number of bits per batch, $y_k$ denotes the true bit value, and $\hat{y}_k$ is the predicted likelihood for bit $k$. This loss function ensures that the model accurately predicts bit likelihoods across all code rates.

The convolutional code decoder and Turbo code decoder are trained separately. Since both decoders are based on the CNE architecture, the convolutional code decoder can share the same CNE0 or CNE1 component with the Turbo code decoder for decoding, but with distinct weights to account for their different coding structures and requirements.
\begin{table}[htbp]
\centering
\caption{Number of Samples in Training, Validation, and Testing Datasets}
\label{tab_datasets}
\begin{adjustbox}{max width=1.0\columnwidth}
\begin{tabular}{ccccccc}
\toprule[0.5pt]
					& \multirow{2.5}{*}{\textbf{Pre-Training}}	& \multirow{2.5}{*}{\textbf{Fine-Tuning}}	& \multirow{2.5}{*}{\textbf{Validation}}	& \multicolumn{2}{c}{\textbf{Testing}} \\
\cmidrule(lr){5-6}
& 																& 											& 											& \textbf{AWGN}		& \textbf{Rayleigh} \\
\midrule[0.5pt]
Convolutional Codes	& $1.6384 \times 10^7$ 						& $1.2288 \times 10^7$ 						& $1.1 \times 10^5$ 						& $1.1 \times 10^6$ & $4.1 \times 10^6$ \\
Turbo Codes 		& $1.6384 \times 10^7$ 						& $1.6384 \times 10^7$ 						& $1.1 \times 10^5$ 						& $1.1 \times 10^6$ & $4.1 \times 10^6$ \\
\bottomrule[0.5pt]
\end{tabular}
\end{adjustbox}
\end{table}

The training, validation, and testing datasets are generated by creating original messages of length $K$, which are then encoded and modulated using BPSK. Noise is added, and the signals are passed through an AWGN channel, following Equation \eqref{eq_awgn_model} to produce the received signals. In the pre-training phase, the dataset contains $128 \times 128 \times 1000 = 1.6384 \times 10^7$ samples. For fine-tuning, datasets include convolutional codes with three code rates and Turbo codes with four code rates, each with $4.096 \times 10^6$ samples. The validation dataset has $1 \times 10^4$ samples per SNR level from 0 to 10 dB. The testing dataset covers two scenarios: AWGN channel with SNRs from 0 to 10 dB and $1 \times 10^5$ samples per SNR, and Rayleigh fading channel with SNRs from 0 to 40 dB and $1 \times 10^5$ samples per SNR. A summary of the datasets is presented in Table~\ref{tab_datasets}. To ensure the independence of training and inference datasets without overlap, different random seeds are used for data generation in these two phases, and a cross-check is performed to ensure that no training data is used for inference.

The training process leverages the Adam optimizer~\cite{adam_opt} with an initial learning rate of $10^{-3}$ during pre-training and $10^{-4}$ during fine-tuning. The reduced learning rate in the fine-tuning phase helps preserve the knowledge acquired during pre-training and minimizes the risk of catastrophic forgetting. To further enhance optimization, a cosine decay scheduler~\cite{consine_sch} is employed to progressively decrease the learning rate to $10^{-6}$ by the end of training. Both pre-training and fine-tuning were conducted for 1000 epochs, with each epoch comprising 128 minibatches of 128 samples each. The training and inference processes are performed on an NVIDIA RTX 4090 GPU with 24GB of memory, complemented by an Intel(R) Xeon(R) Platinum 8468V CPU and 512GB of DDR5 memory, providing the computational power necessary to handle the complexity and diversity of the dataset.

\section{Experiments}
In the experimental setup, the information block length during training is 120, while during inference, it is varied across 120, 240, 480, and 960 to evaluate the model's generalization to different input sizes. Training is performed in an AWGN channel, with inference conducted in both AWGN and Rayleigh fading channels to assess generalization across different channel conditions. Convolutional codes are assigned code rates of 1/2, 2/3, 3/4, and 5/6, in line with protocol specifications. For consistency, the Turbo codes are configured with code rates of 1/3, 1/2, 2/3, 3/4, and 5/6, as Turbo codes offer more flexibility in rate adaptation. The fine-tuning rates are 1/3, 1/2, 2/3, and 3/4, while the 5/6 code rate is used as an unseen rate only during inference to validate the model's generalization ability. For fine-tuning, $\text{SNR}_{\text{offset}}$ is set to 2.5 dB for convolutional decoders and 1.5 dB for Turbo decoders. The traditional convolutional decoder uses the Viterbi algorithm with a traceback depth of 120, while the Turbo code decoder employs a reduced-complexity BCJR algorithm called max-log-MAP~\cite{max_log_map} with full traceback depth. It should be noted that scaling the extrinsic information in the max-log-MAP algorithm can enhance decoding performance~\cite{bcjr_scale}. All simulations of the BCJR algorithm in this paper use the max-log-MAP without a scaling factor.

The Rayleigh fading channel is modeled with 3 independent taps and a MIMO configuration with 4 transmit and 4 receive antennas. Channel state information (CSI) in the Rayleigh fading scenario is obtained using least squares channel estimation~\cite{mimo_chest}. For a received signal  $\boldsymbol{y}$, the transmitted pilot matrix $\bm{\Omega}$, and the channel matrix $\boldsymbol{H}$, the LS estimation is given by:
\begin{equation}
\label{eq_ls_chest}
\boldsymbol{\hat{H}} = \boldsymbol{y}\bm{\Omega}^\dagger
\end{equation}
where $\bm{\Omega}^\dagger = (\bm{\Omega}^H\bm{\Omega})^{-1}\bm{\Omega}^H$ is the Moore-Penrose pseudoinverse of $\bm{\Omega}$, and $\bm{\Omega}^H$ represents the Hermitian transpose of $\bm{\Omega}$. The pilot matrix $\bm{\Omega} \in \mathbb{C}^{N_T \times N_T}$ is generated according to the IEEE 802.11 protocol~\cite{ieee_80211_std} to facilitate accurate channel estimation.

For MIMO detection, the minimum mean square error (MMSE) algorithm is employed~\cite{mimo_sigdet}, which minimizes the mean squared error between the estimated transmitted signal $\hat{\boldsymbol{x}}$ and the true transmitted signal $\boldsymbol{x}$. The MMSE filter $\boldsymbol{W} \in \mathbb{C}^{N_T \times N_R}$ is computed as:
\begin{equation}
\label{eq_mmse_cal_w}
\boldsymbol{W} = \left( \overline{\boldsymbol{H}}^H \overline{\boldsymbol{H}} + \lambda \boldsymbol{I} \right)^{-1} \overline{\boldsymbol{H}}^H
\end{equation}
where $\lambda = 10^{-6}$ is a regularization parameter ensuring numerical stability, and $\overline{\boldsymbol{H}} \in \mathbb{C}^{(N_R + N_T) \times N_T}$ is the extended matrix incorporating the noise variance $\sigma^2$. The extended channel matrix is defined as:
\begin{equation}
\label{eq_mmse_cal_hbar}
\overline{\boldsymbol{H}} = \begin{bmatrix} \boldsymbol{\hat{H}} \\ \sigma^2 \boldsymbol{I} \end{bmatrix}
\end{equation}
where $\boldsymbol{I} \in \mathbb{C}^{N_T \times N_T}$ is the identity matrix.

Once the MMSE filter is computed, the transmitted signal $\boldsymbol{x}$ is estimated as:
\begin{equation}
\label{eq_mmse_cal_x_hat}
\hat{\boldsymbol{x}} = \boldsymbol{W} \boldsymbol{y}
\end{equation}

This process yields the estimated transmitted signal \(\hat{\boldsymbol{x}}\), which is subsequently passed through the demodulation and decoding stages to recover the original message.

Specifically, in the demodulation stage, the LLR for the $i$-th bit of the detected signal is computed based on the Euclidean distance between the detected signal and the possible transmitted symbols. The LLR for the $i$-th bit $x_i$ is calculated as follows:
\begin{equation}
\begin{aligned}
\text{LLR}(x_i) = \log \left( \frac{P(x_i = 0 \mid \hat{x})}{P(x_i = 1 \mid \hat{x})} \right) 
= \min_{\boldsymbol{x}_0} \left( \| \hat{x} - \boldsymbol{x}_0 \| \right) - \min_{\boldsymbol{x}_1} \left( \| \hat{x} - \boldsymbol{x}_1 \| \right)
\end{aligned}
\end{equation}

Here, $\boldsymbol{x}_0$ and $\boldsymbol{x}_1$ represent the possible transmitted symbols corresponding to hypotheses $x_i = 0$ and $x_i = 1$, respectively, while $\hat{x}$ denotes the detected signal.

Additionally, regardless of whether convolutional codes or Turbo codes are employed, to enhance the model's generalization ability across different communication channels, it is essential to normalize the LLRs before feeding them into the neural network, while retaining the sign of the LLRs. The normalization procedure is expressed as follows:
\begin{equation}
	\label{eq_llr_norm}
	\begin{aligned}
		\overline{\mathbf{llr}} = \left\lvert \frac{\mathbf{llr} - \mu}{\sqrt{\delta^2 + \epsilon}} \right\rvert \odot \text{sign}(\boldsymbol{\mathbf{llr}})
	\end{aligned}
\end{equation}
where $\mu$ and $\delta^2$ denote the mean and variance of the LLRs after demodulation but before de-rate matching, comprising both systematic and parity sequences, respectively, and $\epsilon=10^{-6}$ is a small constant added to ensure numerical stability by preventing division by zero. The function $\text{sign}(\cdot)$ preserves the sign of the LLRs during normalization.
\begin{table}[htb]
\centering
\caption{Detailed Parameters for CNE and Simulation}
\includegraphics[width=0.6\columnwidth]{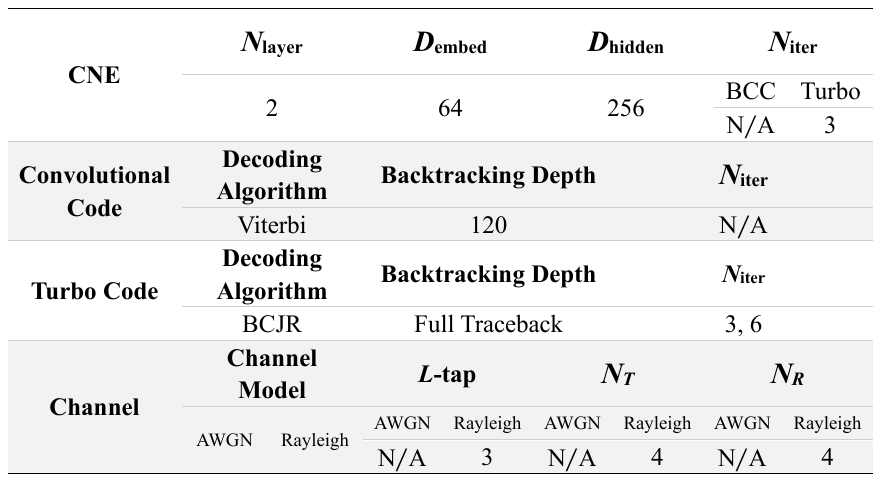}
\label{tab_sim_para}
\end{table}

Following the normalization process, the LLRs are processed by the de-rate matching module and then passed to the proposed CNE neural network, which is characterized by several key hyperparameters. These include the number of LSTM layers $N_{\text{layer}}$, the embedding dimension $D_{\text{embed}}$, the hidden layer size $D_{\text{hidden}}$, and the number of iterations $N_{\text{iter}}$. Specifically, in the LSTM cell shown in Eq.~\ref{eq_lstm}, the weight matrices $\boldsymbol{W}_f, \boldsymbol{W}_i, \boldsymbol{W}_o, \boldsymbol{W}_c \in \mathbb{R}^{D_{\text{embed}} \times (D_{\text{embed}} + D_{\text{hidden}})}$. A detailed summary of the parameters for the CNE neural network architecture and the simulation setup is provided in Table~\ref{tab_sim_para}.

\subsection{AWGN Channels: Benchmarking and Precision}
In this section, we compare the performance of the proposed LSTM-based CNE with traditional decoders in an AWGN channel using BPSK modulation. The information bolck lengths varied from 120 to 960, and the code rates ranged from 1/3 to 5/6. The BER was then measured for each combination of information block length and code rate. For each SNR point, $10^5$ code blocks were simulated to ensure statistical reliability.
\begin{figure*}[htbp]
\centering
\includegraphics[width=1.0\textwidth]{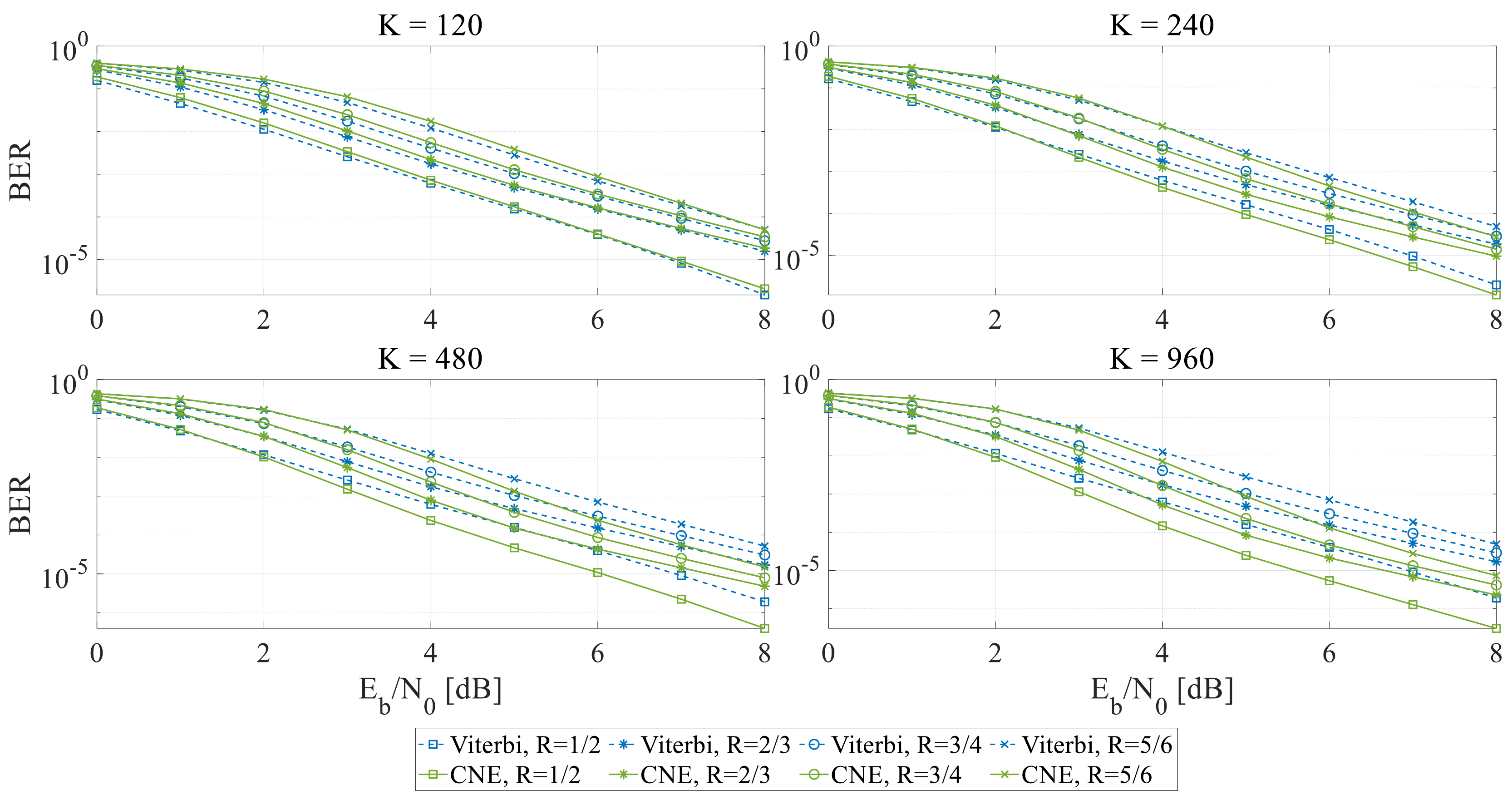}
\caption{Performance comparison of the proposed CNE decoder and the conventional Viterbi algorithm for convolutional codes. The results demonstrate significant performance gains with increasing information block lengths.}
\label{fig_awgn_bcc_sim_result}
\end{figure*}

The simulation results for convolutional codes are depicted in Figure~\ref{fig_awgn_bcc_sim_result}. The proposed decoding approach exhibits comparable performance to the conventional Viterbi algorithm when the information block length is short, such as 120 bits. However, as the block length increases, the proposed CNE decoder significantly outperforms the Viterbi decoder \cite{viterbi_alg_1967}, achieving state-of-the-art (SOTA) performance. Notably, at an information block length of 960 and a BER of $10^{-4}$, the proposed CNE decoder achieves a performance gain of over 1.0 dB across all code rates. This performance advantage stems from the CNE's ability to exploit bidirectional LLR information, capturing both forward and backward dependencies among LLRs. In contrast, the Viterbi algorithm, constrained by limited traceback depth to minimize hardware complexity, fails to achieve a global optimum.
\begin{figure*}[htbp]
\centering
\includegraphics[width=1.0\textwidth]{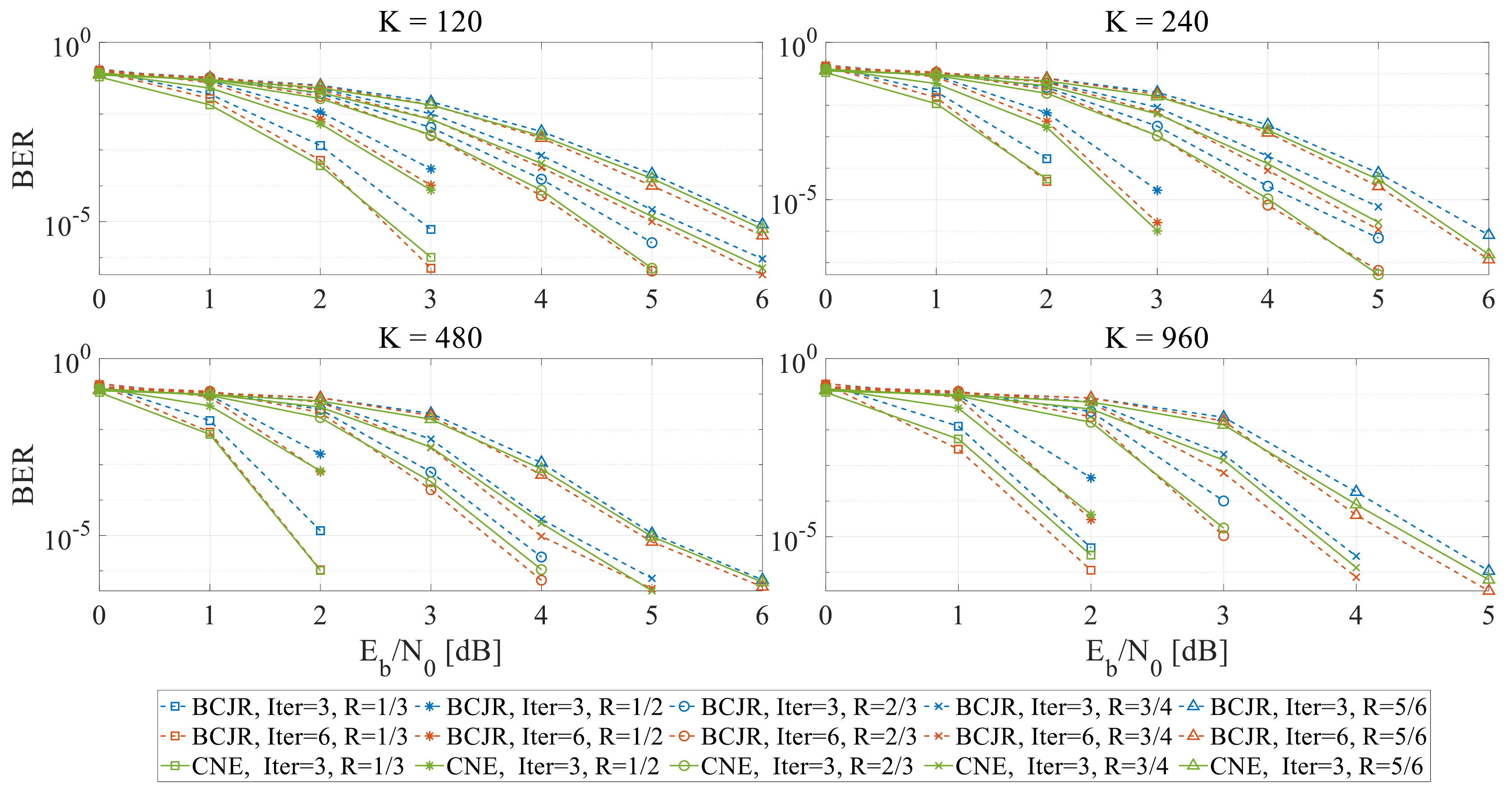}
\caption{BER performance of Turbo codes with the proposed CNE decoder (3 iterations) versus BCJR decoders (3 and 6 iterations without a scaling factor). Results indicate comparable performance with reduced computational complexity.}
\label{fig_awgn_turbo_sim_result}
\end{figure*}

In Figure~\ref{fig_awgn_turbo_sim_result}, the proposed CNE decoder, operating with 3 iterations, achieves performance comparable to the conventional BCJR decoder with 6 iterations. This highlights the CNE's efficiency in significantly reducing computational complexity while preserving high decoding accuracy. Moreover, Figure~\ref{fig_awgn_turbo_sim_result} shows the performance comparison at an identical iteration count of 3. The CNE decoder consistently surpasses the BCJR decoder across all evaluated information block lengths and code rates. This superiority is attributed to the ability of the LSTM-based CNE to effectively model temporal dependencies in the data and refine LLR estimates through its iterative architecture, resulting in enhanced decoding performance.

Notably, due to the explicit embedding of puncturing information into the neural network, the proposed architecture effectively generalizes to the previously unseen 5/6 code rate for both convolutional and Turbo codes, demonstrating its generalization ability across various code rates.

\subsection{Rayleigh Channels: Generalization and Robustness}
In this section, we evaluate the performance of the proposed CNE decoder in a $4\times4$ MIMO Rayleigh fading channel characterized by 3-tap multipath coefficients. The simulation utilized 16-quadrature amplitude modulation (16-QAM) with Gray-coded mapping for signal transmission. The evaluation covered a range of information block lengths from 120 to 960 and code rates from 1/3 to 5/6, mirroring the configurations used in the AWGN scenario. For Turbo codes, the BCJR algorithm performs up to 6 iterations, while the proposed CNE decoder is fixed at 3 iterations. To assess the impact of channel estimation accuracy, we simulated the system's performance under both perfect channel state information and least squares channel estimation. To ensure statistical robustness, $10^5$ code blocks were simulated for each SNR point.

Tables~\ref{tab_rayleigh_bcc_sim_result} and~\ref{tab_rayleigh_turbo_sim_result} present the required $E_b/N_0$ values to achieve target BERs of $10^{-3}$ and $10^{-4}$, respectively, for convolutional and Turbo codes in Rayleigh fading channels. In the tables, $\overline{A}$ indicates that when $E_b/N_0 > A$, the decoding performance converges to a BER of 0 before reaching the target BER, and ``$...$'' signifies that even at $E_b/N_0 = 40$ dB, the target BER remains unattained. Smaller $E_b/N_0$ values reflect better performance, and the best results are highlighted in bold.

From the results in Table~\ref{tab_rayleigh_bcc_sim_result} and~\ref{tab_rayleigh_turbo_sim_result}, it is evident that the proposed CNE decoder achieves a significant performance improvement over traditional decoding algorithms in Rayleigh fading channels. Notably, under LS channel estimation, the CNE decoder demonstrates a substantial advantage, outperforming traditional algorithms by more than 5 dB in many scenarios. Furthermore, the CNE decoder even surpasses traditional decoders with perfect channel state information.

\begin{table}[htbp]
\centering
\caption{Simulation results for convolutional codes in a $4\times4$ Rayleigh fading channel with 16-QAM modulation}
\includegraphics[width=1.0\textwidth]{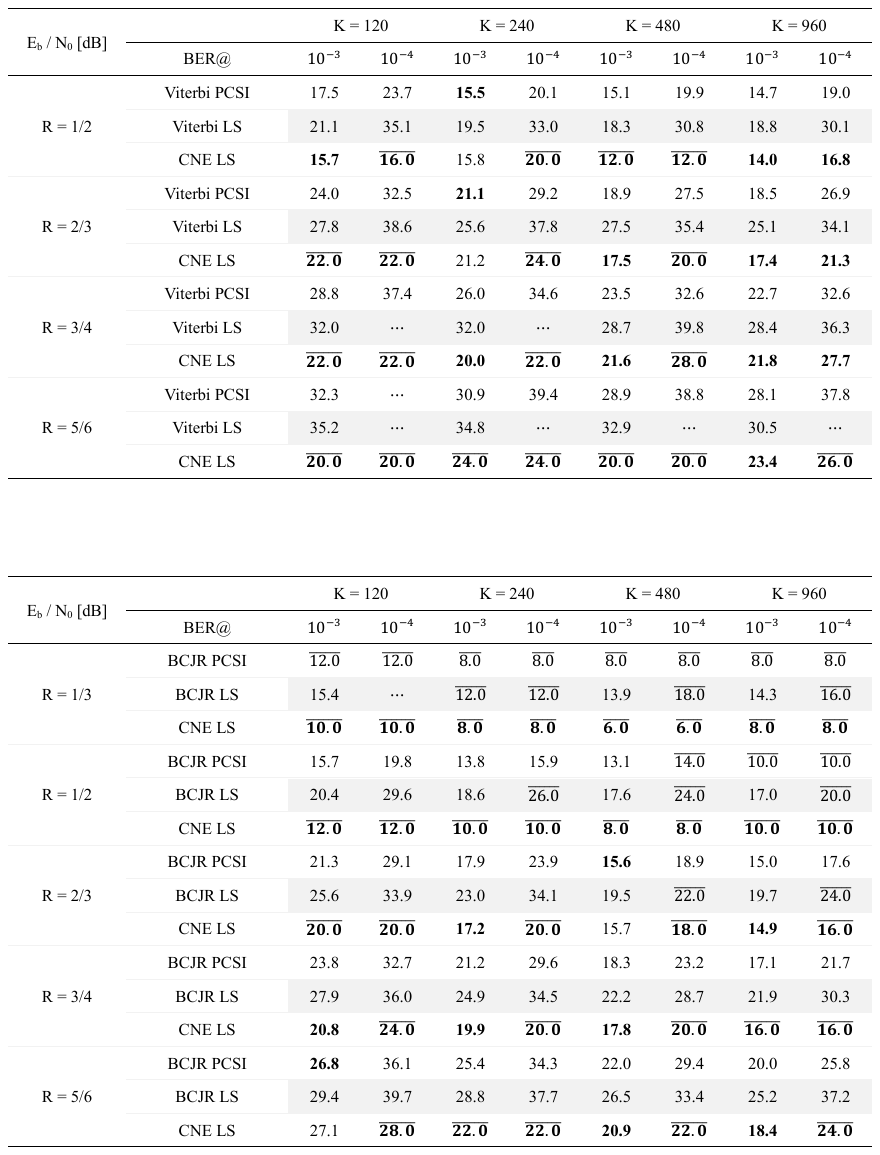}
\label{tab_rayleigh_bcc_sim_result}
\end{table}

\begin{table}[htbp]
\centering
\caption{Simulation results for Turbo codes in a $4\times4$ Rayleigh fading channel with 16-QAM modulation. BCJR decoder without a scaling factor}
\includegraphics[width=1.0\textwidth]{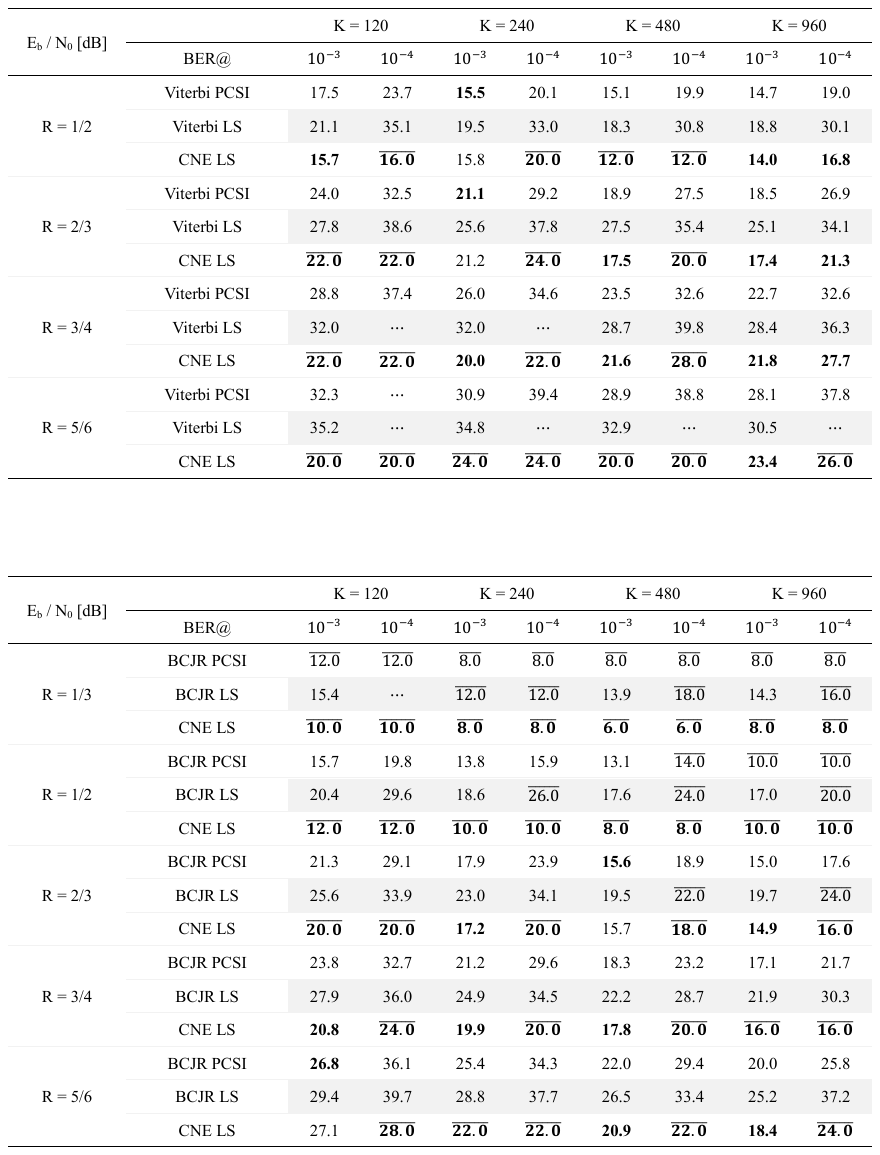}
\label{tab_rayleigh_turbo_sim_result}
\end{table}

The simulation results under Rayleigh fading channels provide critical insights into the strengths of the proposed CNE decoder, particularly in terms of \textbf{generalization} and \textbf{robustness}. The proposed approach achieves consistent performance gains under both AWGN and Rayleigh fading channels. This demonstrates its ability to generalize across different channel conditions, effectively learning the underlying relationships between received LLRs and transmitted bits, irrespective of the noise or fading environment. Unlike traditional decoders that rely on predefined assumptions and fixed traceback mechanisms, the CNE decoder dynamically adapts to complex channel conditions. Its LSTM-based architecture efficiently models temporal dependencies and leverages bidirectional information flow, enabling it to handle multipath interference and Rayleigh fading robustly.

\section{Analysis}
In this section, we examine the proposed CNE decoder, exploring its generalization capabilities, puncturing-aware embedding, computational complexity, decoding latency, and strategies for mitigating complexity and latency in practice. These aspects are crucial for understanding the algorithm's performance, adaptability, and practical applicability in real-world communication systems.

\subsection{Generalization Capabilities}
Following the methodology described in the DeepTurbo paper, we reproduced its implementation (available at \url{https://github.com/TechYan1990/deep-turbo/}) and trained and evaluated it using the simulation parameters specified therein, as shown in Table~\ref{tab_dt_cne_sim_para0}. To ensure a fair comparison, we configured the CNE architecture to match DeepTurbo’s parameters, including a non-shared two-layer bidirectional gated recurrent unit with a hidden dimension $D_{\text{hidden}} = 100$, embedding dimension $D_{\text{embed}} = 5$, and six decoding iterations $N_{\text{iter}} = 6$. Under these conditions, the primary architectural difference between CNE and DeepTurbo lies in the embedding strategy: CNE projects the systematic bits ($\mathbf{llr}_s$) and both parity sequences ($\mathbf{llr}_{z}$, $\mathbf{llr}_{z'}$) into a high-dimensional space, whereas DeepTurbo only projects posterior information.
\begin{table}[htbp]
\centering
\caption{Parameters of DeepTurbo and CNE simulations to analyze generalization capabilities}
\label{tab_dt_cne_sim_para0}
\begin{adjustbox}{max width=0.8\columnwidth}
\huge
\begin{tabular}{ccc}
	\toprule[0.5pt]
											& \textbf{DeepTurbo}         	& \textbf{CNE Turbo}         	\\
	\midrule[0.5pt]
	RNN Architecture             			& Non-Shared 2-Layer Bi-GRU  	& Non-Shared 2-Layer Bi-GRU  	\\
	Training Epochs              			& 1000                       	& 1000                       	\\
	Batch Size                   			& 128                        	& 128                        	\\
	Batches per Epoch            			& 128                        	& 128                        	\\
	Training Information Block Length   	& 100                        	& 100                        	\\
	Inference Information Block Length		& 1000                  		& 1000                       	\\
	Number of Inference Code Blocks 		& 10000                   	 	& 10000                      	\\
	Learning Rate                			& 0.001                      	& 0.001                      	\\
	Optimizer                    			& Adam                       	& Adam                       	\\
	Loss Function                			& BCE							& BCE							\\
	Training SNR                 			& 0 dB                       	& 0 dB                       	\\
	Training Code Rate                   	& $1/3$                      	& $1/3$                      	\\
	Posterior Feature Size $F_\text{s}$ 	& 5                   			& N/A                        	\\
	Embedding Dimension $D_\text{embed}$ 	& N/A                			& 5                          	\\
	Hidden Dimension $D_\text{hidden}$ 		& 100                 			& 100                        	\\
	Number of Iterations $N_\text{iter}$ 	& 6                 	 		& 6                          	\\
	Puncturing-Aware Embedding         		& N/A                       	& None                       	\\
	\bottomrule[0.5pt]
\end{tabular}
\end{adjustbox}
\end{table}
\begin{figure}[htbp]
\centering
\includegraphics[width=0.5\columnwidth]{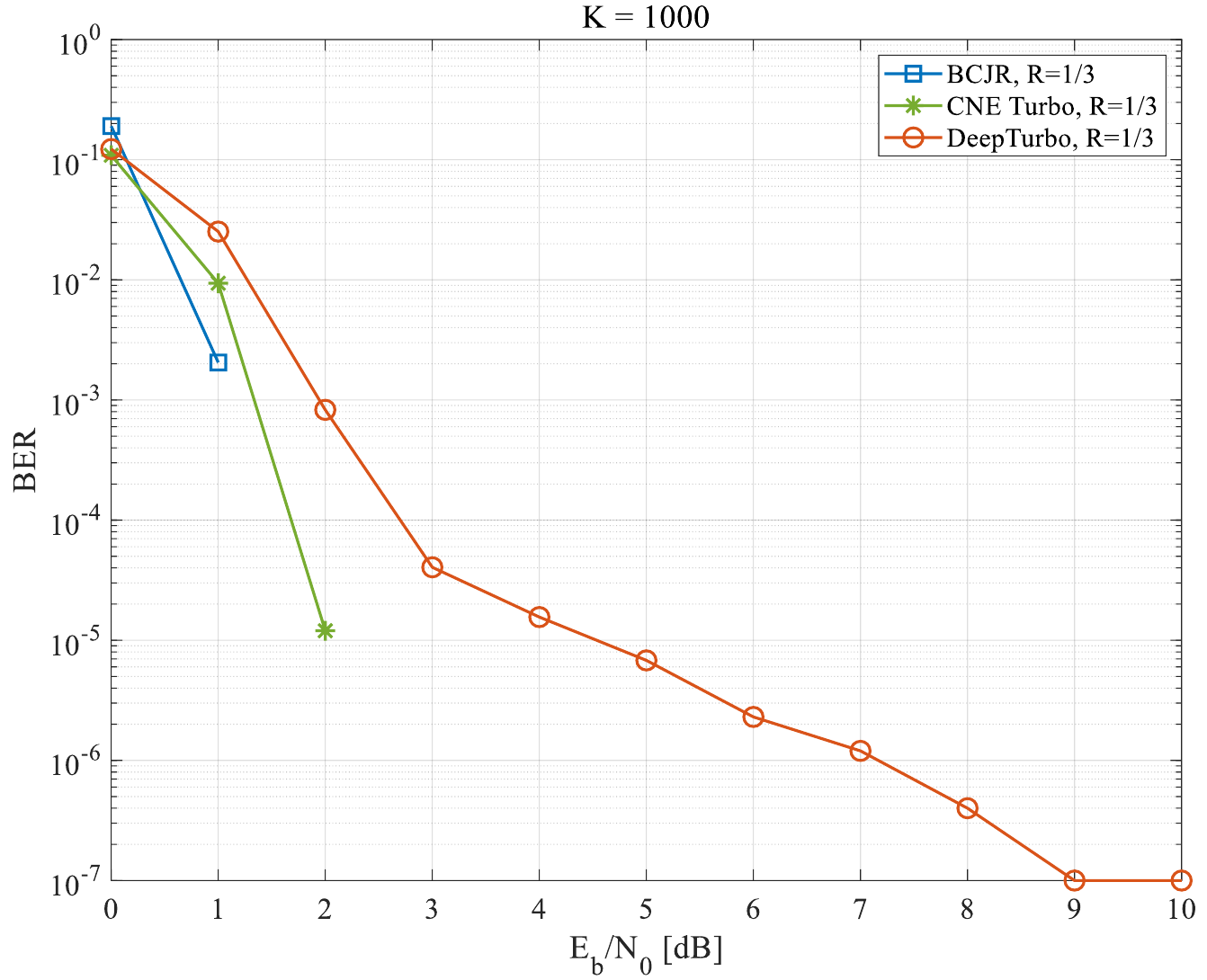}
\caption{BER performance of Turbo codes decoded by CNE, DeepTurbo, and BCJR (6 iterations without a scaling factor) on an AWGN channel with information block length 1000.}
\label{fig_dt_cne_sim_result}
\end{figure}

Both CNE and DeepTurbo were trained on an AWGN channel with a information block length of 100 and evaluated at a block length of 1000 to assess generalization. The BCJR algorithm, used as a baseline, was configured with six iterations. The results, presented in Figure~\ref{fig_dt_cne_sim_result}, demonstrate that DeepTurbo exhibits poor generalization at high SNRs, manifesting as a BER floor, whereas CNE maintains robust performance without this limitation.

These findings indicate that CNE’s superior generalization to different code lengths is not merely a result of increased parameters, as both architectures use identical parameter counts in this experiment. Instead, the key preprocessing step contributing to generalization is CNE’s comprehensive embedding of $\mathbf{llr}_{s}$, $\mathbf{llr}_{z}$, and $\mathbf{llr}_{z'}$ into a high-dimensional space. This approach enhances the neural network’s expressive power, enabling it to capture complex relationships within the codeword more effectively than DeepTurbo’s posterior-only embedding.

\subsection{Puncturing-Aware Embedding}
\begin{table}[htbp]
\centering
\caption{Parameters of DeepTurbo and CNE simulations to analyze puncturing-aware embedding}
\label{tab_dt_cne_sim_para1}
\begin{adjustbox}{max width=0.8\columnwidth}
\huge
\begin{tabular}{ccc}
	\toprule[0.5pt]
											& \textbf{DeepTurbo}         	& \textbf{CNE Turbo}         	\\
	\midrule[0.5pt]
	RNN Architecture             			& Non-Shared 2-Layer Bi-GRU  	& Shared 2-Layer Bi-LSTM  		\\
	Training Epochs              			& 1000                       	& 1000                       	\\
	Batch Size                   			& 128                        	& 128                        	\\
	Batches per Epoch            			& 128                        	& 128                        	\\
	Training Information Block Length   	& 120                        	& 120                        	\\
	Inference Information Block Length		& 120                  			& 120                       	\\
	Number of Inference Code Blocks 		& 10000                   	 	& 10000                      	\\
	Learning Rate                			& 0.001                      	& 0.001                      	\\
	Optimizer                    			& Adam                       	& Adam                       	\\
	Loss Function                			& BCE							& BCE							\\
	Training SNR                 			& 0 dB                       	& 0 dB                       	\\
	Training Code Rate                   	& $1/3$                      	& $1/3$                      	\\
	Fine-Tuning Code Rate                  	& $1/3,1/2,2/3,3/4$             & $1/3,1/2,2/3,3/4$             \\
	Posterior Feature Size $F_\text{s}$ 	& 5                   			& N/A                        	\\
	Embedding Dimension $D_\text{embed}$ 	& N/A                			& 64                          	\\
	Hidden Dimension $D_\text{hidden}$ 		& 256                 			& 256                        	\\
	Number of Iterations $N_\text{iter}$ 	& 6                 	 		& 3                          	\\
	Puncturing-Aware Embedding         		& N/A                       	& None                       	\\
	\bottomrule[0.5pt]
\end{tabular}
\end{adjustbox}
\end{table}
	
\begin{figure}[htbp]
\centering
\includegraphics[width=0.5\columnwidth]{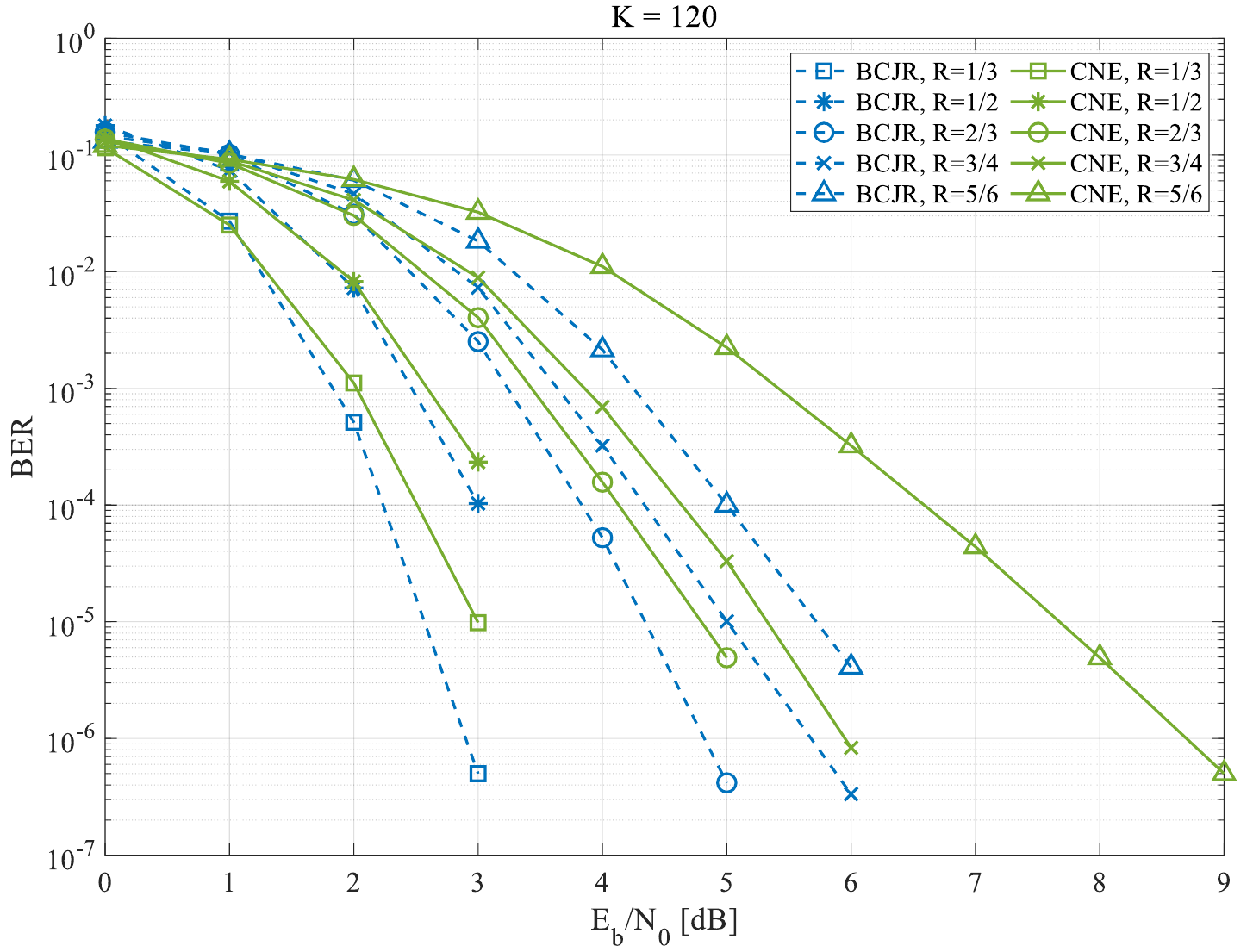}
\caption{BER performance of Turbo codes decoded by CNE (without the puncturing-aware embedding module), and BCJR (6 iterations without a scaling factor) on an AWGN channel with information block length 120.}
\label{fig_bcjr_cne_sim_result}
\end{figure}

\begin{figure}[htbp]
\centering
\includegraphics[width=0.5\columnwidth]{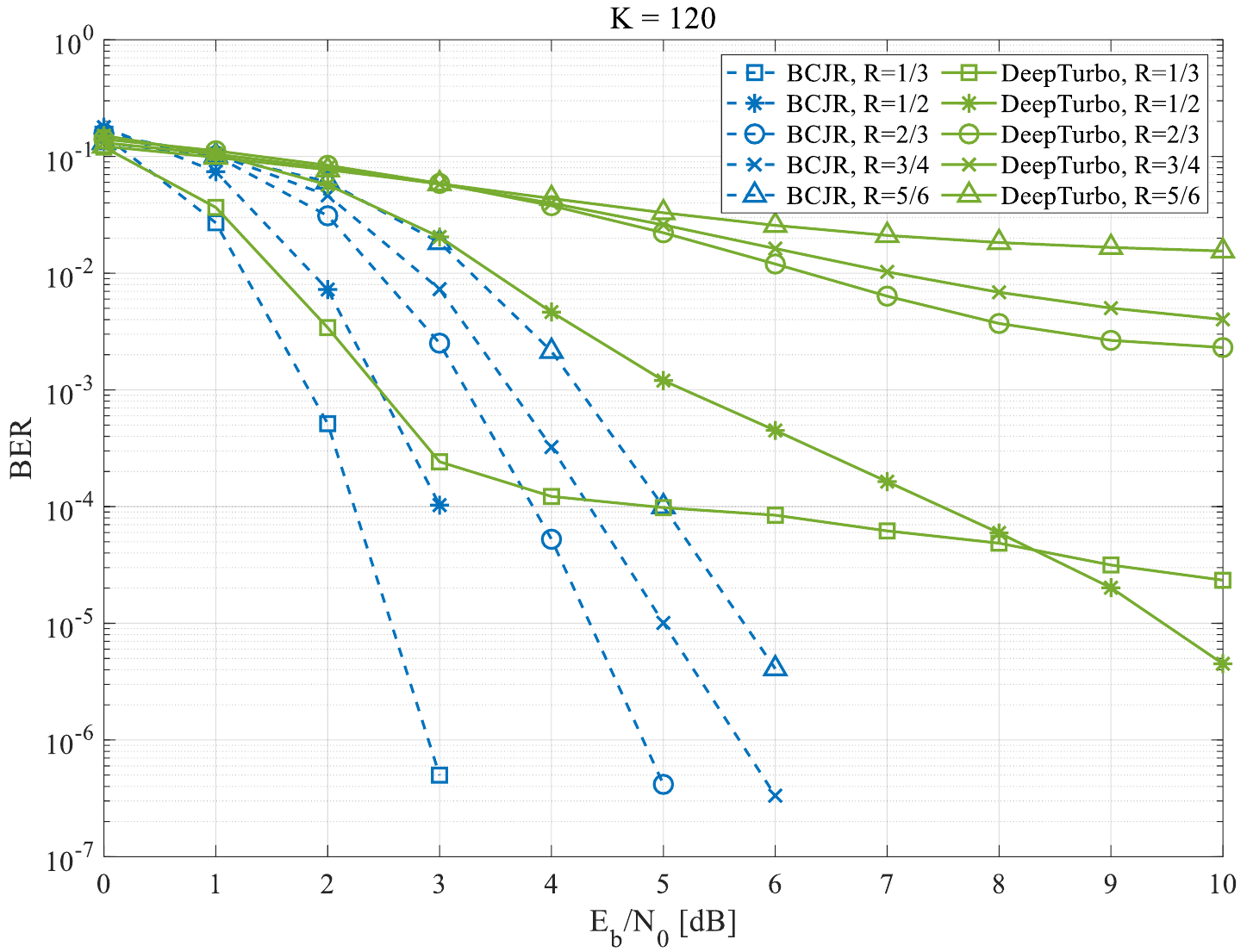}
\caption{BER performance of Turbo codes decoded by DeepTurbo, and BCJR (6 iterations without a scaling factor) on an AWGN channel with information block length 120.}
\label{fig_bcjr_dt_sim_result}
\end{figure}
	
This secion we persent an ablation study on the effect of the puncturing-aware embedding module and the gain of the proposed CNE over a DeepTurbo-like decoder for punctured codes, providing a detailed comparison of decoding performance in terms of BER. This study evaluates CNE (without the puncturing-aware embedding module) against a DeepTurbo-like decoder and the traditional BCJR algorithm under identical training conditions.
	
For a fair comparison, both CNE and DeepTurbo are configured with their respective optimal parameters, where the hyperparameters of DeepTurbo are set according to~\cite{deep_turbo}, and the hidden dimension of DeepTurbo is increased from 100 in the original paper to 256 for a fair comparison, as summarized in Table~\ref{tab_dt_cne_sim_para1}. During fine-tuning, code rates of 1/3, 1/2, 2/3, and 3/4 are used, with the 5/6 code rate reserved as an unseen rate to assess generalization. Training and inference are conducted in an AWGN channel.
\begin{enumerate}
\item \textbf{CNE vs. BCJR}: Figure~\ref{fig_bcjr_cne_sim_result} compares the BER performance of CNE (without the puncturing-aware embedding module) against the BCJR algorithm. When the puncturing-aware embedding module is disabled, CNE exhibits limited generalization to the unseen 5/6 code rate, with a performance degradation of approximately 1.5 dB at BER = $10^{-4}$. For other code rates (1/3, 1/2, 2/3, and 3/4), the BER performance of CNE without the module is comparable to the BCJR algorithm, indicating that the absence of puncturing awareness hampers its ability to adapt to unseen rates.
\item \textbf{DeepTurbo vs. BCJR}: Figure~\ref{fig_bcjr_dt_sim_result} shows the BER performance of the DeepTurbo decoder compared to the BCJR algorithm. DeepTurbo exhibits minimal performance loss for the 1/3 code rate but suffers from a BER floor, indicating limited error correction capability at higher SNRs. For other code rates (1/2, 2/3, 3/4, and 5/6), DeepTurbo shows significant performance degradation and fails to demonstrate meaningful generalization, underscoring its inability to handle punctured codes effectively.
\item \textbf{Key Insight}: From the simulation results in Figure~\ref{fig_bcjr_cne_sim_result}, it is evident that, although the absence of the puncturing-aware embedding module severely degrades the performance of CNE for the 5/6 code rate, it still exhibits some generalization capability. In contrast, the simulation results in Figure~\ref{fig_bcjr_dt_sim_result} show that DeepTurbo, even after fine-tuning, demonstrates no generalization to the 5/6 code rate. This indicates that the generalization ability for code rates is not solely due to the puncturing-aware embedding module but also stems from the contribution of projecting the systematic bits ($\mathbf{llr}_s$) and parity sequences ($\mathbf{llr}_z$, $\mathbf{llr}_{z'}$) into a high-dimensional embedding space.
\end{enumerate}

The ablation study confirms that the puncturing-aware embedding module is a pivotal component of the proposed CNE, enabling superior generalization to unseen code rates (e.g., 5/6) compared to both CNE without the module and a DeepTurbo-like decoder. The module’s ability to encode puncturing patterns into the latent space, combined with high-dimensional embedding of systematic and parity sequences, drives CNE’s robustness and protocol compatibility.
\begin{figure}[htb]
\centering
\includegraphics[width=1.0\columnwidth]{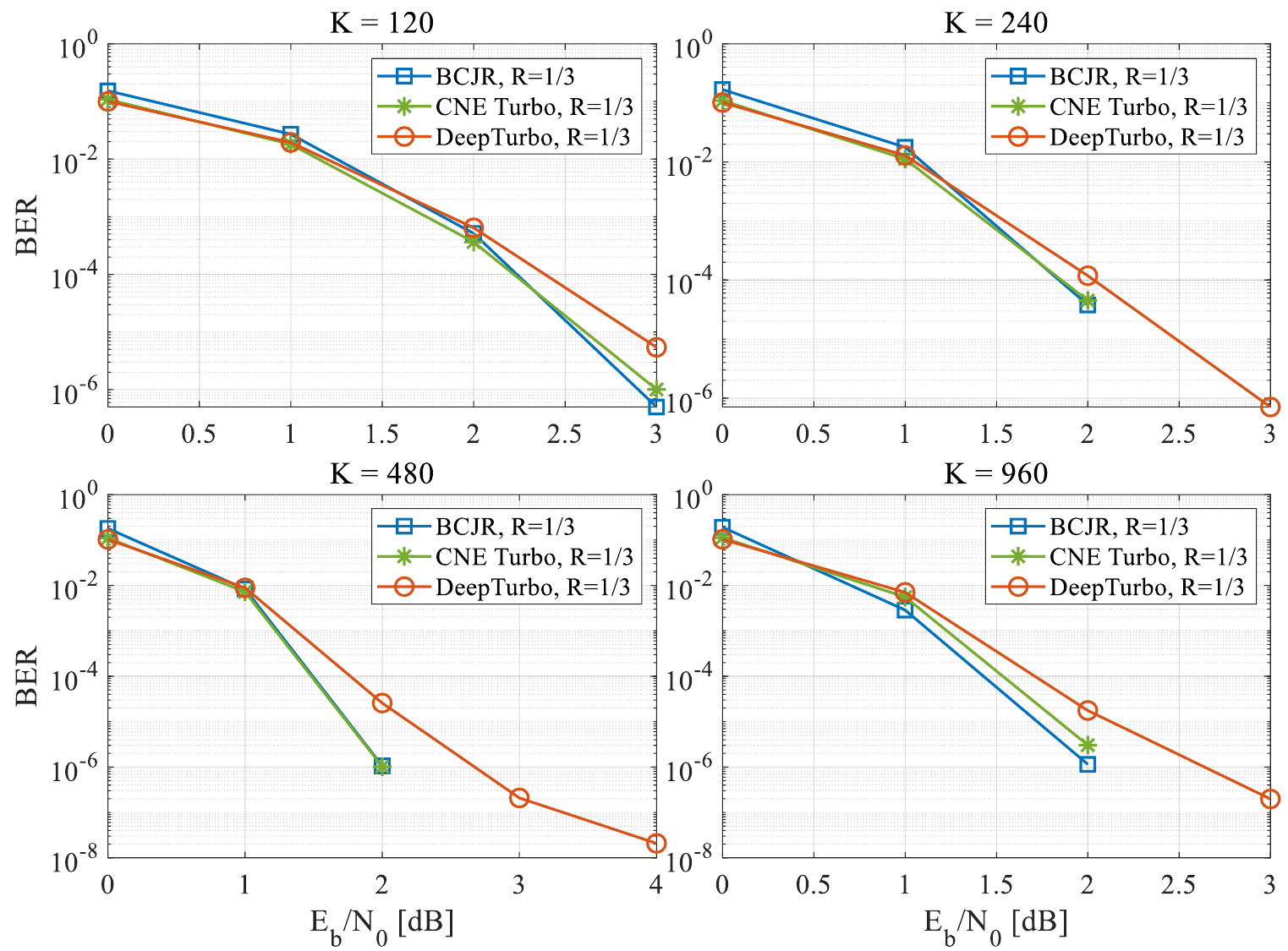}
\caption{BER performance of Turbo codes decoded by CNE, DeepTurbo, and BCJR (6 iterations without a scaling factor) on an AWGN channel with information block lengths of 120, 240, 480, and 960 at a fixed non-punctured code rate of 1/3.}
\label{fig_bcjr_cne_dt_r13_sim_result}
\end{figure}

Based on the analysis above, due to DeepTurbo's limited generalization ability across information block lengths and rates, we trained separate DeepTurbo models for each information block length ($K = 120, 240, 480, 960$) with a fixed non-punctured code rate of 1/3. The hyperparameters were configured according to Table~\ref{tab_dt_cne_sim_para1}, with the fine-tuning step omitted. The simulation results, shown in Figure~\ref{fig_bcjr_cne_dt_r13_sim_result}, indicate that DeepTurbo exhibits a performance gap of approximately 0.2 dB compared to CNE at a BER of $10^{-4}$ under fixed code length and rate conditions.

\subsection{Computational Complexity}
The computational complexity of the proposed CNE decoder, as detailed in Eq. \eqref{eq_lstm_total_process}, arises from four key components: the input projection and gating operations, the batch normalization, the LSTM-based sequence processing, and the output projection.

The input features $\boldsymbol{x} \in \mathbb{R}^{K \times D_{\text{in}}}$ are projected using $\boldsymbol{W}_{\text{proj}} \in \mathbb{R}^{D_{\text{embed}} \times D_{\text{in}}}$, while puncturing patterns $\boldsymbol{p} \in \mathbb{R}^{K \times D_{\text{in}}}$ are processed with $\boldsymbol{W}_{\text{punc}} \in \mathbb{R}^{D_{\text{embed}} \times D_{\text{in}}}$. This step, involving matrix multiplications and sigmoid activations, incurs a complexity of $\mathcal{O}(2K \cdot D_{\text{embed}} \cdot D_{\text{in}} + K \cdot D_{\text{embed}})$.

Following this, batch normalization is applied to the modulated features in $\mathbb{R}^{K \times D_{\text{embed}}}$, with a complexity of $\mathcal{O}(K \cdot D_{\text{embed}})$. The core computational unit, a bidirectional LSTM, processes sequences of length $K$ with input size $D_{\text{embed}}$ and hidden state size $D_{\text{hidden}}$, involving three gates and a candidate memory cell. Its complexity, dominated by quadratic scaling with $D_{\text{hidden}}$, is:
\begin{equation}
\label{eq_lstm_complexity}
\begin{aligned}
\mathcal{O} \left( K \cdot \left( 8 D_{\text{hidden}}^2 + 8 D_{\text{hidden}} D_{\text{embed}} + 14 D_{\text{hidden}} \right) \right)
\end{aligned}
\end{equation}

Finally, the LSTM output $\boldsymbol{h} \in \mathbb{R}^{K \times 2D_{\text{hidden}}}$ is projected using $\boldsymbol{W}_{\text{out}} \in \mathbb{R}^{2D_{\text{hidden}} \times 1}$, with a complexity of $\mathcal{O}(2K \cdot D_{\text{hidden}})$. The total CNE complexity is:
\begin{equation}
\label{eq_cne_total_complexity}
\begin{aligned}
\mathcal{O} \left( K \cdot \left( 
\begin{aligned}
8 D_{\text{hidden}}^2 + 8 D_{\text{hidden}} D_{\text{embed}} +
2 D_{\text{in}} D_{\text{embed}} + 2 D_{\text{embed}} + 16 D_{\text{hidden}}
\end{aligned}
\right) \right)
\end{aligned}
\end{equation}

For comparison, the Viterbi algorithm for a convolutional code with constraint length $L$ has a complexity of:
\begin{equation}
\label{eq_viterbi_complexity}
\begin{aligned}
\mathcal{O}(K \cdot 2^{L-1})
\end{aligned}
\end{equation}

The BCJR algorithm for Turbo decoding, with $N_{\text{iter}}$ iterations, has a complexity of:
\begin{equation}
\label{eq_bcjr_complexity}
\begin{aligned}
\mathcal{O}(N_{\text{iter}} \cdot K \cdot 2^{L+1})
\end{aligned}
\end{equation}

Table~\ref{tab_dt_cne_complexity} compares the complexity of CNE and DeepTurbo decoders, focusing on trainable parameters and multiply-accumulate operations (MACs) per decoded bit, evaluated using Thop~\cite{flops_thop}, a tool designed to measure the MACs of neural networks. Hyperparameters are listed in Table~\ref{tab_dt_cne_sim_para1}, except that CNE includes the puncturing-aware embedding module.
\begin{table}[htbp]
\centering
\caption{Complexity comparison of CNE and DeepTurbo decoder}
\label{tab_dt_cne_complexity}
\begin{adjustbox}{max width=0.7\columnwidth}
\huge
\begin{tabular}{cccc}
	\toprule[0.5pt]
								& Number of Iterations	& Trainable Parameter	& MACs/decoded bit		\\
	\midrule[0.5pt]
	\textbf{CNE Convolutional}	& N/A  					& 2,237,441 			& 2,245,632 			\\
	\textbf{CNE Turbo}      	& 3  					& 6,715,398 			& 16,168,550 			\\
	\textbf{DeepTurbo}      	& 3  					& 9,554,016 			& 11,516,463 			\\
	\textbf{DeepTurbo}     		& 6  					& 19,108,026 			& 23,032,921 			\\
	\bottomrule[0.5pt]
\end{tabular}
\end{adjustbox}
\end{table}

The CNE Turbo decoder, with shared weights between CNE0 and CNE1, significantly reduces trainable parameters compared to DeepTurbo’s non-shared SISO weights. For 3 iterations, DeepTurbo requires approximately 1.42 times the parameters of CNE Turbo, and for 6 iterations, this increases to about 2.85 times. The shared-weight design of CNE enhances efficiency by lowering storage and computational overhead.

In terms of computational complexity, DeepTurbo at 3 iterations requires roughly 0.71 times the MACs per decoded bit of CNE Turbo, due to its lower projection dimension ($F_\text{s} = 5$ vs. CNE’s $D_\text{embed} = 64$). However, at 6 iterations—DeepTurbo’s optimal configuration—it demands approximately 1.42 times the MACs per decoded bit of CNE Turbo. This demonstrates that CNE achieves superior performance with competitive complexity using fewer iterations.

\subsection{Decoding Latency}
Decoding latency is critical for real-time communication systems, and the CNE decoder’s latency stems from input projection and gating, batch normalization, LSTM processing, and output projection.

The input projection maps $\boldsymbol{x} \in \mathbb{R}^{K \times D_{\text{in}}}$ with $\boldsymbol{W}_{\text{proj}} \in \mathbb{R}^{D_{\text{embed}} \times D_{\text{in}}}$, while modulating with $\boldsymbol{p} \in \mathbb{R}^{K \times D_{\text{in}}}$ via $\boldsymbol{W}_{\text{punc}} \in \mathbb{R}^{D_{\text{embed}} \times D_{\text{in}}}$, using parallel matrix-vector multiplications with latency:
\begin{equation}
\label{eq_input_delay}
\begin{aligned}
T_{\text{proj}} = t_{\text{mat}}(D_{\text{embed}}, D_{\text{in}})
\end{aligned}
\end{equation}

Batch normalization then normalizes features in $\mathbb{R}^{K \times D_{\text{embed}}}$ in parallel, with latency:
\begin{equation}
\label{eq_bn_delay}
\begin{aligned}
T_{\text{BN}} = t_{\text{bn}}(D_{\text{embed}})
\end{aligned}
\end{equation}

The sequential LSTM, processing $K$ time steps, dominates latency due to its sequential nature:
\begin{equation}
\label{eq_lstm_delay}
\begin{aligned}
T_{\text{LSTM}} = K \cdot t_{\text{lstm}}(D_{\text{hidden}}, D_{\text{embed}})
\end{aligned}
\end{equation}

The final output projection maps $\boldsymbol{h} \in \mathbb{R}^{K \times 2D_{\text{hidden}}}$ with $\boldsymbol{W}_{\text{out}} \in \mathbb{R}^{2D_{\text{hidden}} \times 1}$, with latency:
\begin{equation}
\label{eq_output_delay}
\begin{aligned}
T_{\text{out}} = t_{\text{mat}}(1, 2D_{\text{hidden}})
\end{aligned}
\end{equation}

The total CNE latency is:
\begin{equation}
\label{eq_cne_total_delay}
\begin{aligned}
T_{\text{CNE}} = T_{\text{proj}} + T_{\text{BN}} + T_{\text{LSTM}} + T_{\text{out}}
\end{aligned}
\end{equation}

The Viterbi algorithm’s latency, scaling with trellis states, is:
\begin{equation}
\label{eq_viterbi_delay}
\begin{aligned}
T_{\text{Viterbi}} = K \cdot t_{\text{state}}(2^{L-1})
\end{aligned}
\end{equation}

The BCJR algorithm, with forward and backward recursions and $N_{\text{iter}}$ iterations, has latency:
\begin{equation}
\label{eq_bcjr_delay}
\begin{aligned}
T_{\text{BCJR}} = 2 \cdot N_{\text{iter}} \cdot K \cdot t_{\text{state}}(2^{L-1})
\end{aligned}
\end{equation}

In addition, we evaluated the proposed CNE decoder using Torch-TensorRT~\cite{tensor_rt}, a tool designed to accelerate neural network inference on NVIDIA GPUs. The evaluation was performed with a floating-point precision of float32, based on 1000 trials with an information block length of 120. The results show that the proposed CNE convolutional code decoder has an average inference latency of $0.116 \, \mu\text{s/decoded bit}$, while the Turbo code decoder exhibits an average inference latency of $0.867 \, \mu\text{s/decoded bit}$.

\subsection{Mitigating Complexity and Latency in Practice}
While the proposed CNE decoder demonstrates superior performance, its computational complexity and latency, particularly due to the sequential nature of the LSTM, pose challenges for real-time and resource-constrained communication systems. To address these drawbacks and ensure practical deployment beyond reliance on GPU acceleration, a multi-faceted approach is presented, encompassing model optimization, hardware acceleration, and scheduling efficiency.
\begin{enumerate}
\item \textbf{Model Optimization}: To reduce computational complexity, advanced compression techniques can be applied. For instance, the 'grow-and-prune' method iteratively removes low-magnitude weights to simplify RNN-based models while preserving accuracy~\cite{grow_and_prune_2020}. Studies on bank-balanced sparsity (BBS) demonstrate 2.3--3.7$\times$ energy savings and 7.0--34.4$\times$ latency reduction for LSTM models on field programmable gate array (FPGA)~\cite{efficient_2019, balanced_2019}. Additionally, techniques such as weight sharing and knowledge distillation compress models effectively, enabling deployment on resource-limited platforms with minimal performance trade-offs~\cite{distilling_2015}.
\item \textbf{Hardware Acceleration}: Beyond GPUs, tailored FPGA-based accelerators offer a promising solution. Pipelined Vector-Scalar Multiplication engines minimize latency through streamlined computation~\cite{sharp_2019}. On-chip static random access memory (SRAM) and embedded dynamic random-access memory (eDRAM) optimize data access, reducing delays critical for real-time systems~\cite{dram_2014, dram_2019}. Furthermore, partial reconfiguration adapts hardware dynamically, balancing performance and power consumption for constrained environments~\cite{compact_cfg_2018}.
\item \textbf{Scheduling Efficiency}: Advanced scheduling strategies enhance latency performance. Unfolded and intergate scheduling parallelize LSTM computations across processing elements, significantly reducing latency for time-sensitive tasks~\cite{serial_scheduling1, serial_scheduling2}. Loop unrolling and pipelining in FPGA designs further eliminate dependencies, ensuring deterministic, low-latency performance~\cite{serial_scheduling3}.
\end{enumerate}

This layered approach—optimizing the model, leveraging specialized hardware, and refining scheduling—effectively mitigates complexity and latency challenges. These strategies ensure the proposed CNE decoder achieves robust, real-time performance, even in resource-constrained settings, making it highly suitable for practical communication systems. Future work could explore specific case studies or quantitative evaluations of these techniques to further validate their effectiveness in diverse scenarios.

\section{Conclusions}
This paper introduces a unified LSTM-based decoding architecture that enhances the performance of punctured convolutional and Turbo codes in practical communication scenarios. By leveraging deep learning techniques, the proposed approach offers a flexible and code-agnostic solution that ensures robust decoding across a wide range of code rates and channel conditions. The results obtained from extensive simulations validate the efficacy of the approach, demonstrating notable improvements in decoding accuracy compared to traditional algorithms. Future work could focus on further optimizing the architecture to reduce decoding complexity and latency, as well as extending its application to more complex coding schemes and diverse communication environments, paving the way for more efficient and reliable decoding in next-generation AI-powered wireless systems.

\bibliographystyle{IEEEtran}
\bibliography{references_final_version}

\end{document}